\documentclass[journal,10pt,onecolumn]{IEEEtran}

\newcommand{\interfrlRewWidth}{0.32}
\newcommand{\interfrlRewSpace}{0cm}

\makeatletter
\def\ps@headings{%
	\def\@oddhead{\mbox{}\scriptsize\rightmark \hfil \thepage}%
	\def\@evenhead{\scriptsize\thepage \hfil \leftmark\mbox{}}%
	\def\@oddfoot{}%
	\def\@evenfoot{}}
\makeatother 

\usepackage[T1]{fontenc}
\usepackage{graphicx}
\usepackage{amsmath, amsfonts,epsfig, multirow, floatflt,float}
\usepackage{amssymb}
\usepackage{diagbox}
\usepackage{cite}
\usepackage{algorithmic}
\usepackage[ruled,lined]{algorithm2e}
\usepackage{hyperref}
\usepackage{enumitem}
\usepackage{mathrsfs}
\usepackage{url}
\usepackage{color}
\usepackage{booktabs}
\usepackage{subcaption}
\usepackage{array}
\usepackage{makecell}

\allowdisplaybreaks

\graphicspath{{figure/}}

\begin{document}
\title{AVDDPG -- Federated reinforcement learning applied to autonomous platoon control}

\author{Christian Boin, Lei Lei, Simon X. Yang}

\maketitle

\begin{abstract}
 Since 2016 federated learning (FL) has been an evolving topic of discussion in the artificial intelligence (AI) research community.  Applications of FL led to the development and study of federated reinforcement learning (FRL).  Few works exist on the topic of FRL applied to autonomous vehicle (AV) platoons. In addition, most FRL works choose a single aggregation method (usually weight or gradient aggregation). We explore FRL's effectiveness as a means to improve AV platooning by designing and implementing an FRL framework atop a custom AV platoon environment.  The application of FRL in AV platooning is studied under two scenarios: (1)  Inter-platoon FRL (Inter-FRL) where FRL is applied to AVs across different platoons; (2) Intra-platoon FRL (Intra-FRL) where FRL is applied to AVs within a single platoon. Both Inter-FRL and Intra-FRL are applied to a custom AV platooning environment using both gradient and weight aggregation to observe the performance effects FRL can have on AV platoons relative to an AV platooning environment trained without FRL.  It is concluded that Intra-FRL using weight aggregation (Intra-FRLWA) provides the best performance for controlling an AV platoon. In addition, we found that weight aggregation in FRL for AV platooning provides increases in performance relative to gradient aggregation. Finally, a performance analysis is conducted for Intra-FRLWA versus a platooning environment without FRL for platoons of length 3, 4 and 5 vehicles. It is concluded that Intra-FRLWA largely out-performs the platooning environment that is trained without FRL.

\end{abstract}

\section{Introduction}
In recent years, federated learning (FL) and its extension federated reinforcement learning (FRL) have become a popular topic of discussion in the artificial intelligence (AI) community.  The concept of FL was first proposed by Google with the development of the federated averaging (FedAvg) aggregation method \cite{McMahan2016FederatedLO}.  FedAvg provided an increase in the performance of distributed systems while also providing privacy advantages when compared to centralized architectures for supervised machine learning (ML) tasks \cite{Konecny2015, BrendanMcMahan2017a, McMahan2016FederatedLO}.  FL's core ideology was initially motivated by the need to train ML models from distributed data sets across mobile devices while minimizing data leakage and network usage \cite{McMahan2016FederatedLO}.

Research on the topics of reinforcement learning (RL) and deep reinforcement learning (DRL) has made great progress over the years; however, there remain important challenges for ensuring the stable performance of DRL algorithms in the real world. DRL processes are often sensitive to small changes in the model space or hyper-parameter space, and as such the application of a single trained model across similar systems often leads to control inaccuracies or instability \cite{Yang2019a, Lim2020}. In order to overcome the stability challenges that DRL poses, often a model must be manually customized to accommodate the finite differences amongst similar agents in a distributed system. FRL aims to overcome the aforementioned issues by allowing agents to share private information in a secure way.  By utilizing an aggregation method, such as FedAvg \cite{McMahan2016FederatedLO}, systems with many agents can have decreased training times with increased performance.

Despite the popularity of FL and FRL, to the best of our knowledge at the time of this study, there are no works applying FRL to platoon control.  In general, there are two types of “models” for AV decision making: vehicle-following modeling and lane-changing modeling \cite{Ye2019}. For the purposes of this study, the vehicle-following approach known as co-operative adaptive cruise control (CACC) is explored. Vehicle following models are based on following a vehicle on a single lane road with respect to a leading vehicle's actions \cite{Zhu2018}.  CACC is a multi-vehicle control strategy where vehicles follow one another in a line known as a platoon, while simultaneously transmitting vehicle data amongst each other \cite{Song2020}.  CACC platoons have been proven to improve traffic flow stability, throughput and safety for occupants \cite{Song2020, Chu2019b}.  Traditionally controlled vehicle following models have limited accuracy, poor generalization from a lack of data, and a lack of adaptive updating \cite{Zhu2018}.

We are motivated by the current state-of-the-art for CACC AV Platoons, along with previous works related to FRL, to apply FRL to the AV platooning problem and observe the performance benefits it may have on the system.  We propose an FRL framework built atop a custom AV platooning environment in order to analyse FRL's suitability for improving AV platoon performance.  In addition, two approaches are proposed for applying FRL amongst AV platoons. The first proposed method is inter-platoon FRL (Inter-FRL), where FRL is applied to AVs across different platoons.  The second proposed method is intra-platoon FRL (Intra-FRL), where FRL is applied to AVs within the same platoon.  We investigate the possibility of Inter-FRL and Intra-FRL as a means to increase performance using two aggregation methods: averaging model weights and averaging gradients. Furthermore, the performance of Inter-FRL and Intra-FRL using both aggregation methods is studied relative to platooning environments trained without FRL (no-FRL). Finally, we compare the performance of Intra-FRL with weight averaging (Intra-FRLWA) against a platooning environment trained without FRL for platoons of length 3, 4 and 5 vehicles.

\subsection{Related works}
In this subsection, the current state-of-the-art is presented for FRL and DRL applied to AV's.  In addition the contributions of this paper are presented.
\subsubsection{Federated reinforcement learning}

There are two main areas of research in FRL currently: horizontal federated reinforcement learning (HFRL), and vertical federated reinforcement learning (VFRL).  HFRL has been selected as the algorithm of choice for the purposes of this study.  HFRL and VFRL differ with respect to the structure of their environments and aggregation methods.  All agents in an HFRL architecture use isolated environments. It follows that each agent's action in an HFRL system has no effect on the other agents in the system.  An HFRL architecture proposes the following training cycle for each agent: first, a training step is performed locally, second, environment specific parameters are uploaded to the aggregation server, and lastly, parameters are aggregated according to the aggregation method and returned to each agent in the system for another local training step. HFRL may be noted to have similarities to “Parallel RL”.  Parallel RL is a long studied field of RL, where agent gradients are transferred amongst each other \cite{Lim2020, Nadiger2019, leilei2021}.

Reinforcement learning is often a sequential learning process, and as such data is often non-IID with a small sample space \cite{sutton2018reinforcement}.  HFRL provides the ability to aggregate experience while increasing the sample efficiency, thus providing more accurate and stable learning  \cite{IntelAI19}. Some of the current works applying HFRL to a variety of applications are summarized below.

A study by Lim \textit{et al.} aims to increase the performance of RL methods applied to multi-IoT device systems.  RL models trained on single devices are often unable to control devices in a similar albeit slightly different environment \cite{Lim2020}.  Currently, multiple devices need to be trained separately using separate RL agents \cite{Lim2020}. The methods proposed by Lim \textit{et al.} sped up the learning process by 1.5 times for a two agent system. In a study by Nadiger \textit{et al.}, the challenges in the personalization of dialogue managers, smart assistants and more are explored.  RL has proven to be successful in practice for personalized experiences; however, long learning times and no sharing of data limit the ability for RL to be applied at scale.  Applying HFRL to atari non-playable characters in pong showed a median improvement of ~17\% for the personalization time \cite{Nadiger2019}. Lastly, Liu \textit{et al.} discuss RL as a promising algorithm for smart navigation systems, with the following challenges: long training times, poor generalization across environments, and storing data over long periods of time \cite{Liu2019b}.  In order to address these problems, Liu \textit{et al.} proposed the architecture `Lifelong FRL', which can be categorized as an HFRL problem.  It is found the Lifelong FRL increased the learning rate for smart navigation system when tested on robots in a cloud robotic system \cite{Liu2019b}.

The successes of the FedAvg algorithm as a means to improve performance and training times for systems have inspired further research into how aggregation methods should be applied.  The design of the aggregation method is crucial in providing performance benefits to that of the base case where FRL is not applied.  The FedAvg \cite{BrendanMcMahan2017a} algorithm proposed the averaging of gradients in the aggregation method.  In contrast, Liang \textit{et al.} proposed using model weights in the aggregation method for AV steering control \cite{Liang2019}. Thus, FRL applications can differ based upon the selection of which parameter to use in the aggregation method.  A study by Zhang \textit{et al.} explores applying FRL to a decentralized DRL system optimizing cellular vehicle-to-everything communication\cite{ZhangX2020}. Zhang \textit{et al.} utilize  model weights in the aggregation method, and describe a weighting factor dividing the sum batch size for all agents by the training batch size for a specific agent\cite{ZhangX2020}.  In addition, the works of Lim \textit{et al.} explore how FRL using gradient aggregation can improve convergence speed and performance on the OpenAI-gym environments CartPole-V0, MountainvehicleContinuous-V0, Pendulum-V0 and Acrobot-V1 \cite{LimHyun2021}. Lim \textit{et al.} determined that aggregating gradients using FRL creates high performing agents for each of the OpenAI-gym environments relative to models trained without FRL\cite{LimHyun2021}.  In addition, Wang \textit{et al.} apply FRL to heterogeneous edge caching \cite{WangXiaofei2021}.  Wang \textit{et al.} show the effectiveness of FRL using weight aggregation to improve hit rate, reduce average delays in the network and offload traffic\cite{WangXiaofei2021}. Lastly, Huang \textit{et al.} apply FRL using model weight aggregation to Service Function Chains in network function virtualization enabled networks\cite{Huang2021}. Huang \textit{et al.} observe that FRL using model weight aggregation provides benefits to convergence speed, average reward and average resource consumption\cite{Huang2021}.

Despite the differences in FRL applications within the aforementioned studies, each study maintains a similar goal: to improve the performance of each agent within the system.  None of the aforementioned works explore the differences in whether gradient or model weight aggregation is favourable in performance, and many of the works apply FRL to distributed network or communications environments.  It is the goal of this study to conclude whether model weight or gradient aggregation is favourable for AV platooning, as well as be one of the first (if not the first) to apply FRL to AV platooning.

\subsubsection{Deep reinforcement learning applied to AV platooning} \label{sec:avRL}
In recent years, there has been a surge in autonomous vehicle (AV) research, likely due to the technologies potential for increasing road safety, traffic throughput and fuel economy \cite{Makantasis2020a, Ye2019}. Two areas of research are often considered when delving into an AV model: supervised learning or RL \cite{Makantasis2020a}.  Driving is considered a multi-agent interaction problem, and due to the large variability of road data, it can be quite challenging (or near impossible) to gather a data set variable enough to train a supervised model \cite{ElSallab2017a}.  Driving data is collected from humans, which can also limit an AI's ability to that of human level \cite{Ye2019}.  In contrast, RL methods are known to generalize quite well \cite{Makantasis2020a}. RL approaches are model-free and a model may be inferred by the algorithm while training.

In order to improve the limitations of vehicle following models, DRL has been a steady area of research in the AV community, with many authors contributing works to DRL applied to CACC \cite{Lin2019, Song2020, Chu2019b, Peake2020}.  In a study by Lin \textit{et al.}, a DRL framework is designed to control a CACC AV platoon\cite{Lin2019}.  The DRL framework uses the deep deterministic policy gradient (DDPG) \cite{Lillicrap2016} algorithm and is found to have near-optimal performance \cite{Lin2019}. In addition, Peake \textit{et al.} identify limitations in platooning with regard to the communication in platooning\cite{Peake2020}.  Through the application of a multi-agent reinforcement learning process, i.e. a policy gradient RL and LSTM network, the performance of a platoon containing 3-5 vehicles is improved upon that of current RL applications to platooning \cite{Peake2020}. Furthermore, Model Predictive Control (MPC) is the current state-of-the-art for real-time optimal control practices \cite{Lin_2021}.  The study performed by Lin \textit{et al.} applies both MPC and DRL methodologies to the AV platoon problem, observing a DRL model trained using the DDPG algorithm produces merely a 5.8\% episodic cost higher than the current state-of-the-art\cite{Lin_2021}. The works of Yan \textit{et al.} propose a hybrid approach to the AV platooning problem where the platoon is modeled as a Markov Decision Process (MDP) in order to collect two rewards from the system at each time step simultaneously\cite{yan2021hybrid}.  This approach also incorporates jerk, the rate of change of acceleration in the calculation of the reward for each vehicle in order to ensure passenger comfort \cite{yan2021hybrid}.  The hybrid strategy led to increased performance to that of the base DDPG algorithm, as the proposed framework switches between using classic CACC modeling and DDPG depending on the performance degradation of the DDPG algorithm \cite{yan2021hybrid}. In another study by Zhu \textit{et al.}, a DRL model is formulated and trained using DDPG to be evaluated against real world driving data. Parameters such as time to collision, headway, and jerk were considered in the DRL model's reward function\cite{Zhu2019}.  The DDPG algorithm provided favourable performance to that of the analysed human driving data, with regard to more efficient driving via reduced vehicle headways, and improved passenger comfort with lower magnitudes of jerk\cite{Zhu2019}. As Vehicle-to-Everything (V2X) communications are envisioned to have a beneficial impact on the performance of platoon controllers, the works of Lei \textit{et al.} investigates the value of V2X communications for DRL-based platoon controllers.  Lei \textit{et al.} emphasizes the trade-off between the gain of including exogenous information in the system state for reducing uncertainty and the performance erosion due to the curse-of-dimensionality\cite{leileiDRL2}.

When formulating the AV platooning problem as a DRL model DDPG is prominently selected as the algorithm for training.  DDPG's ability to handle continuous actions space and complex state's is perfect for the CACC platoon problem.  However, despite the DDPG algorithm's success in literature, there are still instability challenges related to the algorithm along with a time consuming hyper-parameter tuning process to account for the minute differences in vehicle models/dynamics amongst platoons. As previously discussed, FRL provides advantages in these areas where information sharing can accelerate performance during training and improve the performance of the system as a whole. In addition, the ability to share experience across like models has been proven to allow for fast convergence of models, which further optimizes the performance of DDPG when applied to AV platoons \cite{Lim2020}.

\subsection{Contributions}
To the best of our knowledge, no works at the time of this study existed covering the specific topic of FRL applied to platoon control. Many of the works existing on FRL have shown the benefits of FRL with regard to the increased rate of convergence and overall system performance with distributed networks, edge caching and communications \cite{ZhangX2020, LimHyun2021, WangXiaofei2021, Huang2021}.  Furthermore, of the works cited in this study, the works closely related to FRL for platoon control are those of Peake \textit{et al.} and Liang \textit{et al.}\cite{Peake2020, Liang2019}.  In contrast to Liang \textit{et al.}, where FedAvg is applied successfully to control the steering angle of a single vehicle, we apply FRL to an AV platooning problem where the control of multiple vehicles' positions and spacing are required \cite{Liang2019}.  Peake \textit{et al.} explore multi-agent reinforcement learning and its ability to improve the performance of AV platoons experiencing communication delays\cite{Peake2020}.  Although Peake \textit{et al.} are also successful in their approach, there is no specific reference to FRL in the paper\cite{Peake2020}. In addition, a variety of existing works on FRL choose to use either gradients or model weights in the FRL aggregation method.  This study explores how both aggregation methods can provide benefits to the AV platooning problem and, most importantly, which provides a better result. Finally, this study further distinguishes its approach from existing literature by declaring two possible ways to apply FRL to AV platooning:

\begin{enumerate}
    \item Intra-FRL: where multi-vehicle platoons share data during training to increase the performance of vehicles within the same platoon.
    \item Inter-FRL: where multi-vehicle platoons share data during training across platoons amongst vehicles in the exact same platoon position to increase performance.
\end{enumerate}

In contrast to existing literature, where it is common to average the parameters across each model in the system, for Intra-FRL, we propose a directional averaging where follower vehicles incorporate the preceding vehicle parameters in the computation of the gradients or weights. Thus, in Intra-FRL, the leading vehicle trains independently of those following. The AV platoon provides a unique playground environment suitable for exploring the suitability of FRL as a means to increase the performance of systems with regard to convergence rate and performance.

\section{Proposed Framework} \label{sec:propsol}

In this section, a state space model is formulated and presented for the AV platooning problem. Next, the MDP model is presented, outlining the platoon system's state space, action space and reward function. Lastly, the FRL DDPG algorithm design and application to AV platooning are described.
\subsection{CACC CTHP model formulation}
Consider a platoon $P$ of vehicles $\mathcal{V}={V_1,V_2,...,V_n}$ where the leader of the platoon is $V_1$.

\begin{figure}[!t]
    \centering
    \includegraphics[width=0.8\linewidth]{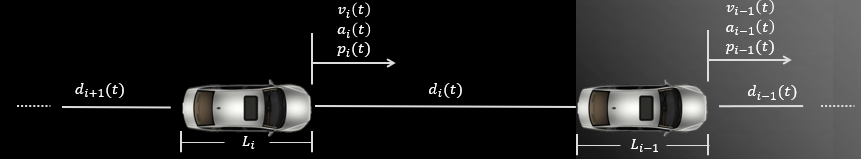}
    \caption{An example platoon modeled with system parameters.}
    \label{fig:platoonmodel}
\end{figure}
\noindent As illustrated in Figure \ref{fig:platoonmodel}, for a general vehicle ($V_i$), the position of $V_i$'s front bumper is defined as $p_i$. The velocity, acceleration and control input of $V_i$ are denoted as $v_i$, $a_i$ and $u_i$.  Furthermore, the acceleration of $V_i$'s predecessor may be denoted as $a_{i-1}$. The control input for $V_i$ is defined as $u_i$ (whether $V_i$ should accelerate or decelerate).  $V_i$'s drive-train dynamics coefficient is defined as $\tau_i$, where large values of $\tau_i$ indicate larger response times for a given input $u_i$ to generate acceleration $a_i$. Lastly, the length of $V_i$ is denoted as $L_i$.  The system dynamics for $V_i$ are thus provided below as

\begin{equation}
    \begin{aligned}\label{eqn:systemparams}
        \dot{p_i(t)} &= v_i(t) \\
        \dot{v_i(t)} &= a_i(t) \\
        \dot{a_i(t)} &= -\frac{1}{\tau_i}a_i(t) + \frac{1}{\tau_i}u_i(t) \\
        \dot{a_{i-1}(t)} &= -\frac{1}{\tau_{i-1}}a_{i-1}(t) + \frac{1}{\tau_{i-1}}u_{i-1}(t)
    \end{aligned}
\end{equation}
The headway $d_i(t)$ in a CACC model is the positional difference of the current vehicle relative to the rear bumper of its leader, which can be derived as \cite{Lin2019, leileiDRL}
\begin{equation}
 \label{eqn:headway}
 d_i(t) = p_{i-1}(t) - p_i(t) - L_{i-1}.
\end{equation}
In addition, the desired headway $d_{r,i}(t)$ is defined as

\begin{equation}
    d_{r,i}(t) = r_i + h_iv_i(t), \label{eqn:deshead}
\end{equation}
where $r_i$ is the standstill distance, and $h_i$ is the time-gap for $V_i$ to maintain relative to it's predecessor $V_{i-1}$.  The position error $e_{pi}$ and the velocity error $e_{vi}$ are defined as:
\begin{equation}
\begin{aligned}
    e_{pi}(t) &= d_i(t) - d_{r,i}(t) \\
    e_{vi}(t) &= v_{i-1}(t) - v_i(t) \\
\end{aligned}
\end{equation}
Therefore, the state of $V_i$ can be defined as $x_i(t) = \begin{bmatrix}e_{pi}(t) & e_{vi}(t) & a_i(t) & a_{i-1}(t)\end{bmatrix}^\top$, and the derivative of the state is:
\begin{equation}
\begin{aligned}
    \dot{e_{pi}(t)} &= e_{vi}(t) - h_ia_i(t),\\
    \dot{e_{vi}(t)} &= a_{i-1}(t) - a_i(t), \\
    \dot{a_i(t)} &= -\frac{1}{\tau_i}a_i(t) + \frac{1}{\tau_i}u_i(t), \\
    \dot{a_{i-1}(t)} &= -\frac{1}{\tau_{i-1}}a_{i-1}(t) + \frac{1}{\tau_{i-1}}u_{i-1}(t).\\
\end{aligned}
\end{equation}
The state space formula for $V_i$ is thus given as
\begin{equation}
    \dot{x_i(t)} = A_ix_i(t) + B_iu_i(t) + C_iu_{i-1}(t), \label{eqn:stspace}
\end{equation}
where $A_i$, $B_i$, and $C_i$ are defined below as

\begin{equation} \label{eqn:systemmatrices}
    \begin{aligned}
        A_{i} &= \begin{bmatrix}
            0 & 1 & -h_i & 0 \\
            0 & 0 & -1 & 1 \\
            0 & 0 & -\frac{1}{\tau_i} & 0 \\
            0 & 0 & 0 & -\frac{1}{\tau_{i-1}}
        \end{bmatrix} \qquad
        B_{i} = \begin{bmatrix}
            0 \\ 0 \\ \dfrac{1}{\tau_i} \\ 0
        \end{bmatrix} \qquad
        C_{i} &= \begin{bmatrix}
            0 \\ 0 \\ 0 \\ \frac{1}{\tau_{i-1}}
        \end{bmatrix}.
    \end{aligned}
\end{equation}

\subsection{MDP model formulation}
The AV platooning problem can be formulated as an MDP problem, where the optimization objective is to minimize the previously defined $e_{pi}$, $e_{vi}$, $u_i$ and lastly jerk.
\subsubsection{2.2.1. State space}
The state space formula \eqref{eqn:stspace} can be discretized using the forward euler method giving the system equation below

\begin{equation}
    \begin{split} \label{eqn:modelBStateSpace}
        x_{i,k+1} &= A_{Di}x_{i,k} + B_{Di}u_{i,k} + C_{Di}u_{i-1, k}, \\
    \end{split}
\end{equation}

\noindent where $x_{i,k}=[e_{pi,k}, e_{vi,k}, a_{i,k}, a_{i-1,k}]$ is the observation state for the MDP problem that includes the position error $e_{pi,k}$, velocity error $e_{vi,k}$, acceleration $a_{i,k}$, and the acceleration of the predecessor vehicle $a_{i-1,k}$ at time step $k$. Moreover, $A_{Di}$, $B_{Di}$, and $C_{Di}$ are given as

\begin{equation} \label{eqn:eulerMatr}
    \begin{aligned}
        A_{Di} &= \begin{bmatrix}
            1 & T & -Th_i & 0\\
            0 & 1 & -T & T\\
            0 & 0 & -\dfrac{T}{\tau_i} + 1 & 0 \\
            0 & 0 & 0 & -\dfrac{T}{\tau_{i-1}} + 1
        \end{bmatrix} \qquad
        B_{Di} = \begin{bmatrix}
            0 \\ 0 \\ \dfrac{T}{\tau_i} \\ 0
        \end{bmatrix} \qquad
        C_{Di} &= \begin{bmatrix}
            0 \\ 0 \\ 0 \\ \frac{T}{\tau_{i-1}}
        \end{bmatrix} .
    \end{aligned}
\end{equation}

\subsubsection{Action space}
Each vehicle within a single lane platoon follows the vehicle in front of it, and as such the only action the vehicle may take to maintain a desired headway is to accelerate, or decelerate. The action for the system is defined as the control input $u_{i,k}$ to the vehicle.

\subsubsection{Reward function}
The design of a reward in a DDPG system is critical to providing good performance within the system.  In the considered driving scenario, it is logical to minimize position error, velocity error, the amount of time spent accelerating and the jerkiness of the driving motion.  The proposed reward thus includes the normalized position error, $e_{pi,k}$, velocity error $e_{vi,k}$, control input $u_{i,k}$ and lastly the jerk. The vehicle reward $c_{i,k}$ is given below, where $a$ $b$, $c$ and $d$ are system hyperparameters.
\begin{equation} \label{eqn:rewardB}
\small
\begin{aligned}
    c_{i,k} &= -\left(a\frac{|e_{pi,k}|}{\max(e_{pi,k})} + b\frac{|e_{vi,k}|}{\max(e_{vi,k})} + c\frac{|u_{i,k}|}{\max(u_{i,k})} +  d\frac{\dot{|a_{i,k}|}}{2\max(a_{i,k})}\right) \\
\end{aligned}
\end{equation}

\subsection{FRL DDPG algorithm}
In this section, the design for implementing the FRL DDPG algorithm on the AV platooning problem is presented.
\subsubsection{DDPG model description}
\begin{figure}[!t]
    \centering
    \includegraphics[width=0.6\linewidth]{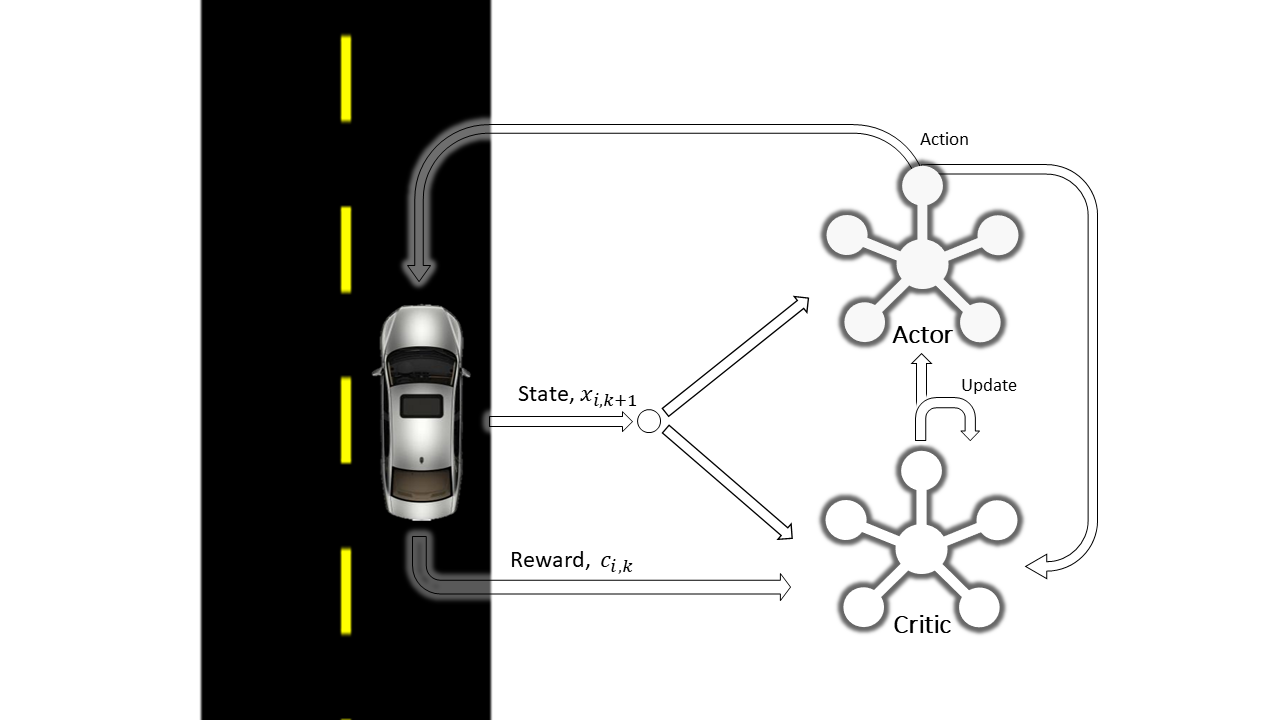}
    \caption{High level flow diagram of the DDPG model for a general vehicle $V_i$ in a platoon.}
    \label{fig:ddpgdraw}
\end{figure}

The DDPG algorithm is composed of an actor, $\mu$ and a critic, $Q$. The actor produces actions $u_t \in \mathbf{U}$ given some observation $x_t \in \mathbf{X}$ and the critic makes judgements on those actions while training using the Bellman equation \cite{Lillicrap2016, sutton2018reinforcement}. The actor is updated by the policy gradient \cite{Lillicrap2016}.  The critic network uses its weights $\theta^q$ to approximate the optimal action-value function $Q(x, u|\theta^q)$ \cite{Lillicrap2016}.  The actor network uses weights $\theta^\mu$ to represent the agents' current policy $\mu(x|\theta^\mu)$ for the action-value function \cite{Lillicrap2016}.  The actor $\mu(x): \mathbf{X} \xrightarrow{} \mathbf{U}$ maps the observation to the action.  Experience replay is used to mitigate the issue of training samples not being independent and identically distributed due to their generation from sequential explorations \cite{Lillicrap2016}.  Two additional models, the target actor $\mu'$ and critic $Q'$ are used in DDPG to stabilize the training of the actor and critic networks by updating parameters slowly based on the target update coefficient $\tau$.  A sufficient value of $\tau$ is chosen such that stable training of $\mu$ and $Q$ is observed.  Figure \ref{fig:ddpgdraw} provides a high level simplified overview of how the DDPG algorithm interacts with a single vehicle in a platoon.

\subsubsection{Inter and intra FRL}
Modifications to the base DDPG algorithm are needed in order to implement Inter-FRL and Intra-FRL.  In order to implement FedAvg the following modifications are required:

\begin{enumerate}
    \item An FRL server: responsible for averaging the system parameters for use in a global update
    \item Model weight aggregation: storing of each model's weights for use in aggregation
    \item Model gradient aggregation: storing of each model's gradients for use in aggregation
\end{enumerate}

In order to perform FRL, it has been proven that including an update delay between global FRL updates is beneficial for performance \cite{Lim2020}. In addition, turning off FRL partway through training is important to allow each agent to refine their models independently of each other such that they can perform best with respect to their environments \cite{Lim2020}.  Lastly, it has also been shown that global updates and local updates should not be performed in the same episode \cite{Liang2019}.

\begin{figure}[!t]
    \centering
    \includegraphics[width=0.69\linewidth]{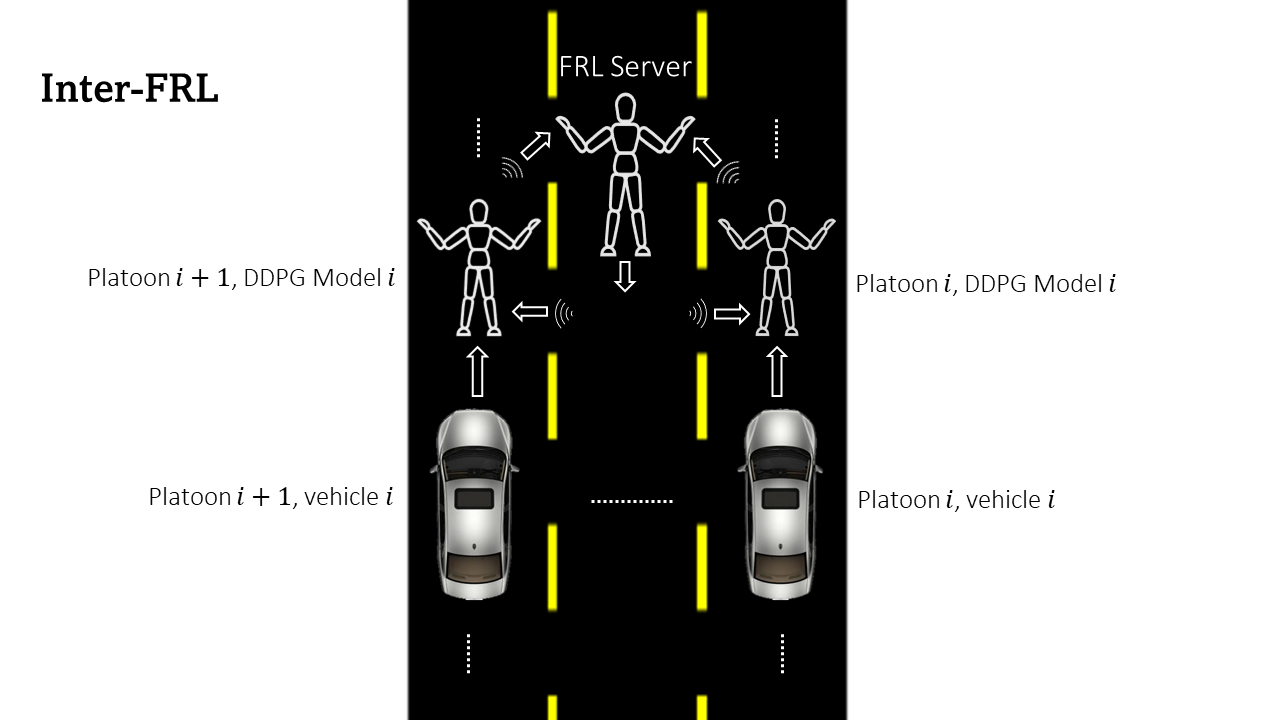}
    \caption{Inter-FRL.}
    \label{fig:interfrl}
\end{figure}

\begin{figure}[!t]
    \centering
    \includegraphics[width=0.69\linewidth]{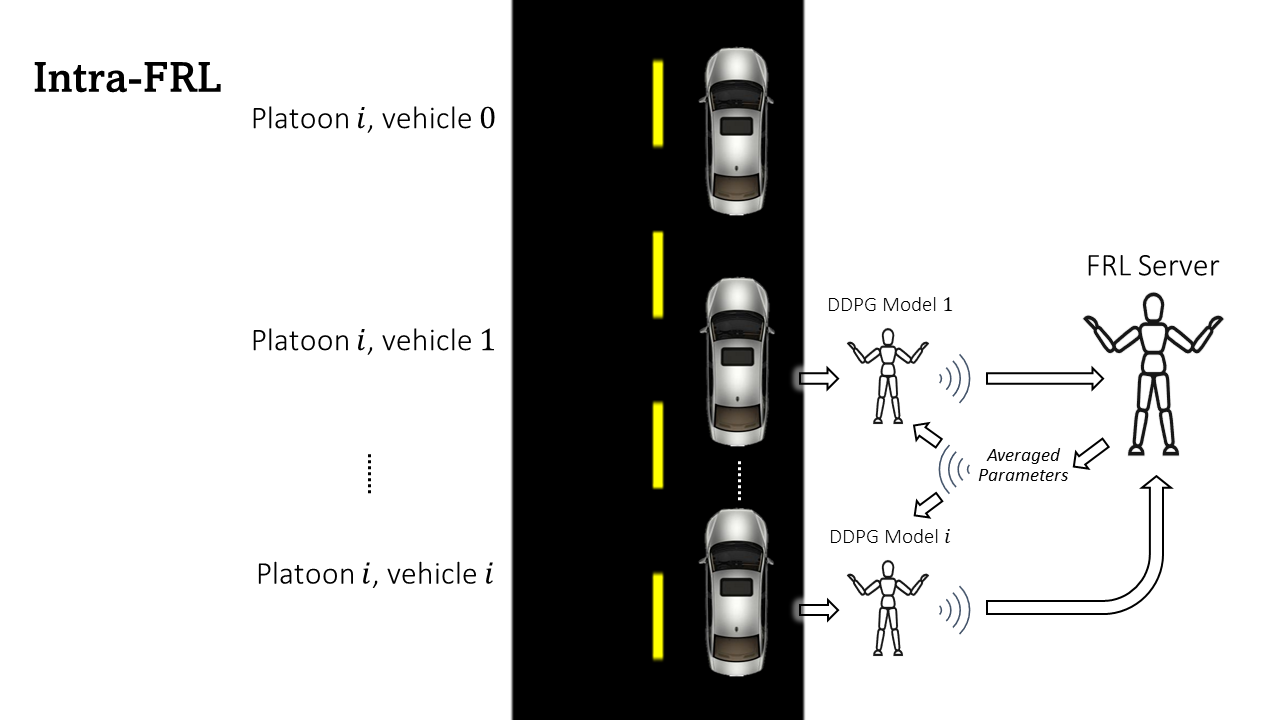}
    \caption{Intra-FRL.}
    \label{fig:intrafrl}
\end{figure}

Two methods of aggregation are implemented in the system design, Inter-FRL (see Figure \ref{fig:interfrl}), and Intra-FRL (see Figure \ref{fig:intrafrl}).  The proposed system is capable of aggregating both the model weights and gradients for each model so that either type of parameter may be averaged for use in global updates.  The FRL server has the responsibility of averaging the parameters (model weights or gradients) across each agent in the system.


The pseudo-code for the Inter/Intra-FRL algorithm is presented in Algorithm 1. The system is designed to allow the training of any number of equal length platoons.  At the lowest level, a DDPG agent exists for each vehicle in each platoon. As such, a DRL model must be initialized for each vehicle in the whole system.  Each DDPG agent trains separately from the others before data is uploaded to the FRL server. Federated averaging is applied at a given time delay known as the FRL update delay, while being terminated at a given episode as defined by the cutoff ratio as seen in Table \ref{tab:frlhyps}. Currently, Algorithm 1 is synchronous, and the FRL server is also synchronous.

\IncMargin{1em}
\begin{algorithm}
\small
\For{each platoon $p \in platoons$}{
   \For{$v\in vehicles$}{
       initialize replay buffer $R_{i}$\;
       initialize actor $\mu_{i}$, critic $Q_{i}$, target actor $\mu'_{i}$, target critic $Q'_{i}$\;
   }
}
\BlankLine
\For{$episode \in training\_episodes$}{
   \For{$p \in platoons$}{
       collect all vehicles states $x_{i,k}$ from $p$\;
   }
   \For{$step \in steps\_per\_episode$}{
       \For{$p\in platoons$}{
           \For{$v\in vehicles$}{
               collect actions $u_{i,k}$ from actor\;
           }
           advance the platoon $p$, with $u_{i,k}$\;
           collect $(x_{i,k}, x_{i,k+1}, c_{i,k}, terminal)$ from $p$\;
       }
       \For{$p\in platoons$}{
           \For{$v\in vehicles$}{
               add $(x_{i,k}, x_{i,k+1}, c_{i,k}, terminal)$ to replay buffer $R_{i}$\;
               \If{FRL update is not required}{
                   train $\mu_{i}$, $Q_{i}$, $\mu'_{i}$, $Q'_{i}$ locally\;
               }
               append gradients of $\mu_{i}$ and $Q_{i}$ to all\_gradients\;
               append weights of $\mu_{i}$ and $Q_{i}$ to all\_weights\;
           }
       }
       \If{FRL update required}{
           \If {gradient averaging enabled} {
               avg\_gradients $\xleftarrow{}$ global\_update(all\_gradients)\;
               train $\mu_{i}$, $Q_{i}$ using avg gradients\;
           }
           \If {weight averaging enabled} {
               avg\_weights $\xleftarrow{}$ global\_update(all\_weights)\;
               update weights $\mu_{i}$, $Q_{i}$, $\mu'_{i}$, $Q'_{i}$ using avg weights\;
           }
       }
   }
}
\BlankLine
\SetKwProg{Fn}{Function}{ is}{end}
\Fn{global\_update(params)}{
   upload params to FRL server\;
   collect averaged params from FRL server\;
   return averaged params\;
}
\caption{FRL applied to an AV platoon.}
\end{algorithm}\DecMargin{1em}

\section{Experimental Results} \label{sec:experiment}
In this section, the experimental setup for applying both Inter and Intra-FRL to the AV platooning environment is presented.  The AV platooning environment and Inter/Intra FRL algorithms are implemented in Python 3.7 using Tensorflow 2.

\subsection{Experimental setup}
The parameters specific to the AV platoon environment are summarized in Table \ref{tab:pl_hyps}.  The time step interval is $T=0.1 s$, and each training episode is composed of 600 time steps.  Furthermore, the coefficients $a$, $b$, $c$ and $d$ given in the reward function \eqref{eqn:rewardB} are a means to define how much each component of \eqref{eqn:rewardB} contributes to the calculation of the reward. These coefficients may be tuned in order to
determine a balance amongst each component, leading to better optimization during training.
The coefficients were tuned using a grid search strategy and are listed as $a=0.4, b = c = d = 0.2$.

\begin{table}[!t]
    \centering
    \caption{Parameters of the AV platoon environment.}
    \begin{tabular}{lll} \toprule
        \textbf{Parameter} & \textbf{Value} \\ \midrule
        Time step $T$ interval & 0.1 s \\
        Number of time steps per training episode & 600 \\
        Time gap $h_i$ &              1 s \\
        Driveline dynamics coefficient $\tau$ &            0.1 s \\
        Maximum absolute control input $u_{max}$ & 2.5 $m/s^2$ \\
        Reward coefficient $a$ &            0.4 \\
        Reward coefficient $b$ &            0.2 \\
        Reward coefficient $c$ &            0.2 \\
        Reward coefficient $d$ &            0.2 \\ \bottomrule
    \end{tabular}
    \label{tab:pl_hyps}
\end{table}

\begin{figure*}[!t]
\centering
\begin{subfigure}{.4\textwidth}
    \centering
    \includegraphics[width=0.75\linewidth]{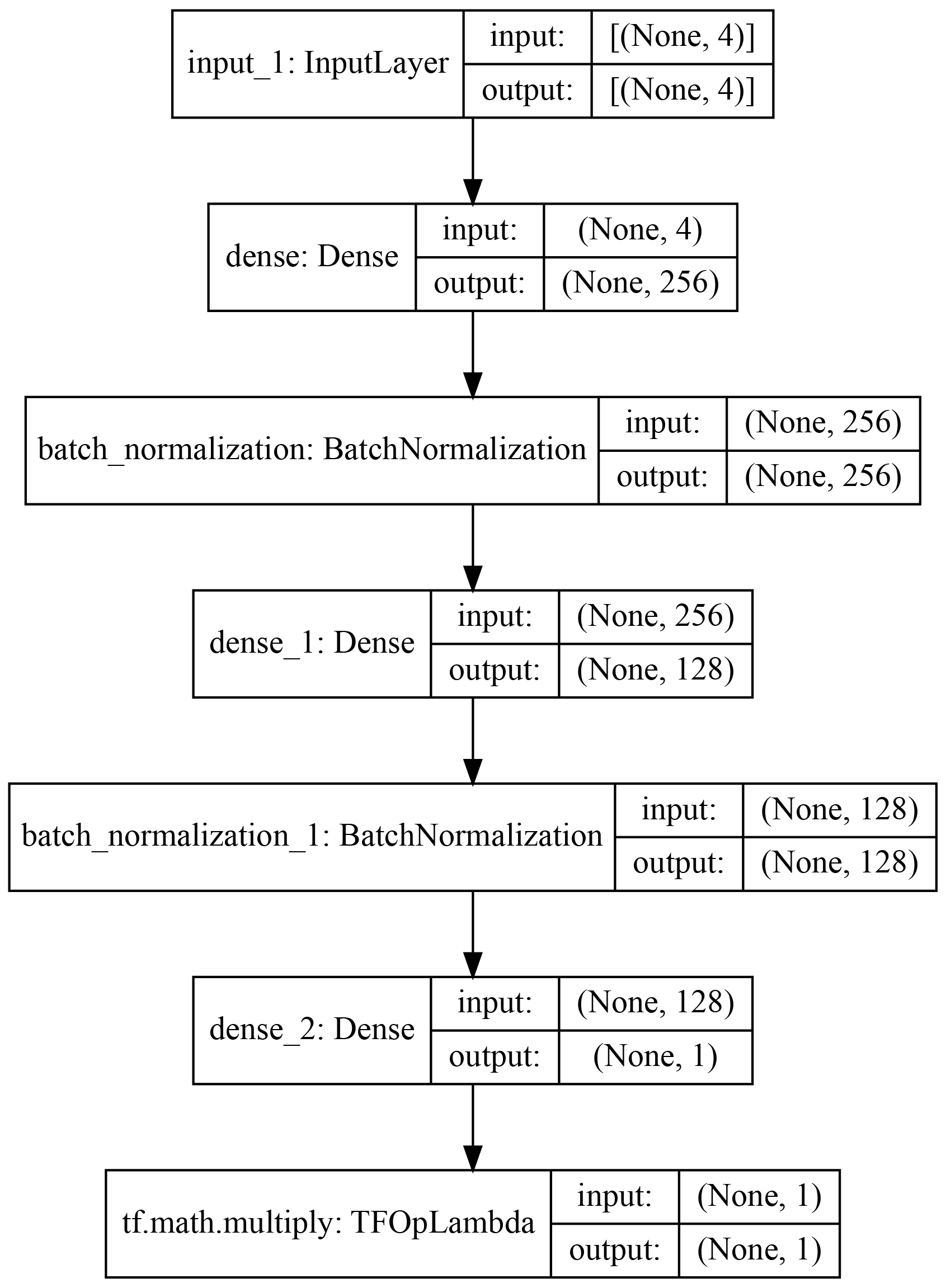}
    \caption{The actor network for $V_i$.}
    \label{fig:actB}
\end{subfigure}%
\begin{subfigure}{.6\textwidth}
    \centering
    \includegraphics[width=1.0\linewidth]{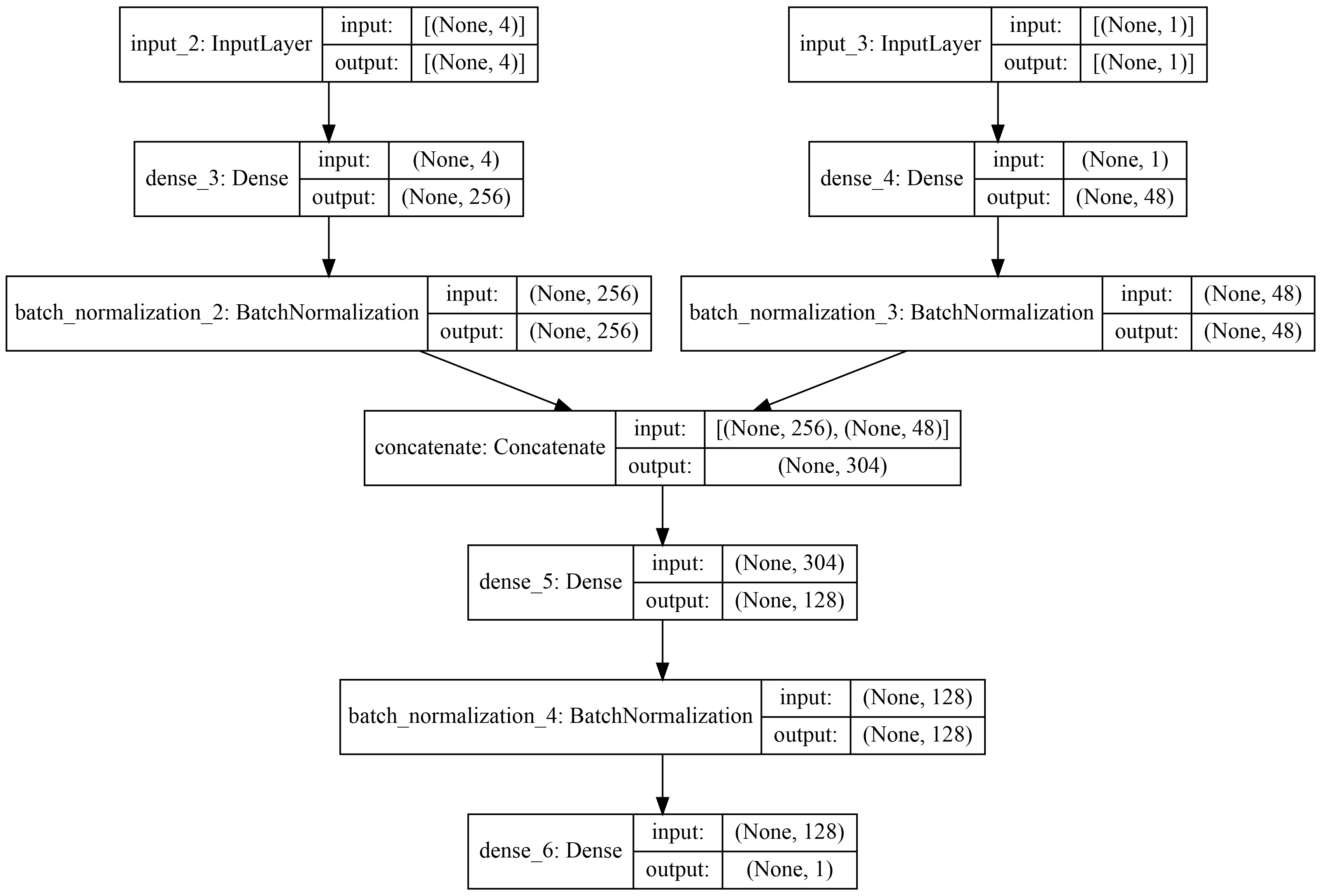}
    \caption{The critic network for $V_i$.}
    \label{fig:crticB}
\end{subfigure}
\caption{Actor and critic networks for $V_i$.}
\label{fig:test}
\end{figure*}

Each DDPG agent consists of a replay buffer, and networks for the actor, target actor, critic and target critic.  The actor network contains four layers: an input layer for the state, two hidden layers with 256 and 128 nodes, respectively, and an output layer. Both hidden layers use batch normalization and the relu activation function.  The output layer uses the tanh() activation function.  The output layer is scaled by the high bound for the control output, in this case 2.5 $m/s^2$.  The critic network is structured with two separate input layers for state and action.  These two layers are concatenated together, and fed into a single hidden layer before the output layer.  The layer with the state input has 48 nodes, the relu activation function and batch normalization.  The same is applied for the action layer, but instead with 256 nodes.  The post concatenation layer uses 304 input nodes, followed by a hidden layer with 128 nodes, again with relu activation and batch normalization applied.  The output of the critic uses a linear activation function.  Ornstein-Uhlenbeck noise is applied to the model's predicted action, $u_i$. The structure of the models is presented in Figure \ref{fig:actB} and \ref{fig:crticB}. All except the final layers of the actor and critic networks were initialized within the range \begin{tiny}$\left[-\dfrac{1}{\sqrt{fan\:in}},\:\dfrac{1}{\sqrt{fan\:in}}\right]$\end{tiny}, where-as the final layer is initialized using a random uniform distribution bounded by $[-3\times10^{-3},\:3\times10^{-3}]$.  Table \ref{tab:ddpg_hyps} presents the hyperparameters specific to the DDPG algorithm.

\begin{table*}[!t]
    \centering
    \caption{Hyperparameters for the DDPG algorithm.}
    \begin{tabular}{lll} \toprule
        \textbf{Hyperparameter} & \textbf{Value} \\ \midrule
        Actor learning rate & 5e-05\\
        Critic learning rate & 0.0005\\
        Batch size & 64\\
        Noise & Ornstein-Uhlenbeck Process with $\theta=0.15$, $\sigma=0.02$ \\
        Weights and Biases  & random uniform distribution $[-3\times10^{-3},\:3\times10^{-3}]$ (final layer), \\
        Initialization    & \begin{tiny}$\left[-\dfrac{1}{\sqrt{fan\:in}},\:-\dfrac{1}{\sqrt{fan\:in}}\right]$\end{tiny} (other layers)\\ \bottomrule
    \label{tab:ddpg_hyps}
    \end{tabular}
\end{table*}

The hyperparameters specific to Inter and Intra-FRL are presented in Table \ref{tab:frlhyps}. During a training session with FRL, both local updates and FRL updates with
aggregated parameters are applied to each DDPG agent in the system. FRL updates usually
occur at a given frequency known as the FRL update delay, and furthermore, FRL updates
may be terminated at a specific training episode as defined by the FRL cutoff ratio.  The FRL update delay is defined as the time in seconds between FRL updates during a training episode. The FRL cutoff ratio is the ratio of the number of episodes where FRL updates are applied divided by the total number of episodes in a training session. Note that the aggregation method denotes whether the model gradients or weights are averaged during training using FRL.
\begin{table}[!t]
    \caption{FRL specific initial hyperparameters.}
    \centering
    \begin{tabular}{llll} \toprule

        \textbf{FRL type} & \textbf{Aggregation method} & \textbf{Hyperparmeter} & \textbf{Value} \\ \midrule

        Inter-FRL & Gradients & FRL update delay &  0.1 \\
        Inter-FRL & Gradients & FRL cutoff ratio &  0.8  \\
        Inter-FRL & Weights & FRL update delay &  30  \\
        Inter-FRL & Weights & FRL cutoff ratio &  1.0   \\
        Intra-FRL & Gradients & FRL update delay &  0.4  \\
        Intra-FRL & Gradients & FRL cutoff ratio &  0.5  \\
        Intra-FRL & Weights & FRL update delay &  0.1  \\
        Intra-FRL & Weights & FRL cutoff ratio &  1.0  \\ \bottomrule

    \end{tabular}
    \label{tab:frlhyps}
\end{table}

For the purposes of this study, an experiment is defined as a training session for a specific configuration of hyper-parameters, using the algorithm defined in Algorithm 1. During each experiment training session, model parameters were trained through the base DDPG algorithm or FRL in accordance with Algorithm 1. Once training has concluded, a simulation is performed using a custom built evaluator API.  The evaluator performs simulations for a single 60 second episode using the trained models, calculating the cumulative reward of the model(s) in the experiment.  The entire project is designed and implemented using Python3, and Tensorflow.  As previously stated, each vehicle in the platoon is modelled using the CACC CTHP model described in Section 3.  For the purposes of this study, multiple sets of DRL experiments were conducted, using 4 random seeds (1-4) for training and a single random seed (6) across all evaluations.

\subsection{Inter-FRL}
In order to evaluate the effectiveness of Inter-FRL relative to the base case where a DRL model is trained using DDPG without FRL, 4 experiments are conducted without Inter-FRL (no-FRL), and 8 with.  For each of the 12 conducted experiments, 2 platoons with 2 vehicles each were trained using one of the four random seeds.  Once training across the four seeds has completed, the cumulative reward for a single evaluation episode is evaluated.  For the experiments using Inter-FRL, two aggregation methods are examined.  First, the gradients of each model are averaged during training, and second, the model weights are averaged.  The multi platoon system trains and shares the aggregated parameters (gradients or weights) amongst vehicles with the same index across platoons. The federated server is responsible for performing the averaging, and each vehicle performs a training episode with the averaged parameters in addition to their local training episodes in accordance with the FRL update delay and FRL cutoff ratio (see Table \ref{tab:frlhyps}). Note that here-after Inter-FRL with gradient aggregation is denoted Inter-FRLGA, and Inter-FRL with weight aggregation is denoted Inter-FRLWA.

\subsubsection{Performance across 4 random seeds}
The performance for each of the systems is calculated by averaging the cumulative reward of each vehicle in the 2 vehicle 2 platoon system, as summarized in Table \ref{tab:interfrl-summary}.  For each of the 3 cases (base case, Inter-FRLGA and Inter-FRLWA), training sessions were run using 4 random seeds.  In order to determine the highest performing system overall, an average and standard deviation is obtained from the result of training using the 4 random seeds. From Table \ref{tab:interfrl-summary}, it is observed that both Inter-FRL scenarios using gradient and weight aggregation provide large performance increases to that of the base case.

\begin{table*}[!t]
  \centering
  \scriptsize
  \caption{Performance after training across 4 random seeds. Each simulation result contains 600 time steps.}
    \begin{tabular}{lrrlrrr} \toprule
    \textbf{Training method} & \multicolumn{1}{l}{\textbf{Seed 1}} & \multicolumn{1}{l}{\textbf{Seed 2}} & \multicolumn{1}{l}{\textbf{Seed 3}} & \multicolumn{1}{l}{\textbf{Seed 4}} & \multicolumn{1}{l}{\textbf{Average system reward}} & \multicolumn{1}{l}{\textbf{Standard deviation}} \\ \midrule
    No-FRL & -3.73 & -2.89 & -4.69 & -3.38 & -3.67 & 0.66 \\
    Inter-FRLGA & -2.79 & -2.81 & -3.05 & -2.76 & -2.85 & 0.11 \\
    Inter-FRLWA & -2.64 & -2.88 & -2.92 & -2.93 & -2.84 & 0.12 \\ \bottomrule
    \end{tabular}%
  \label{tab:interfrl-summary}%
\end{table*}%

 \subsubsection{Convergence properties}
\begin{figure}[!t]
\centering
    \begin{subfigure}{0.3\textwidth}
        \includegraphics[width=0.99\textwidth]{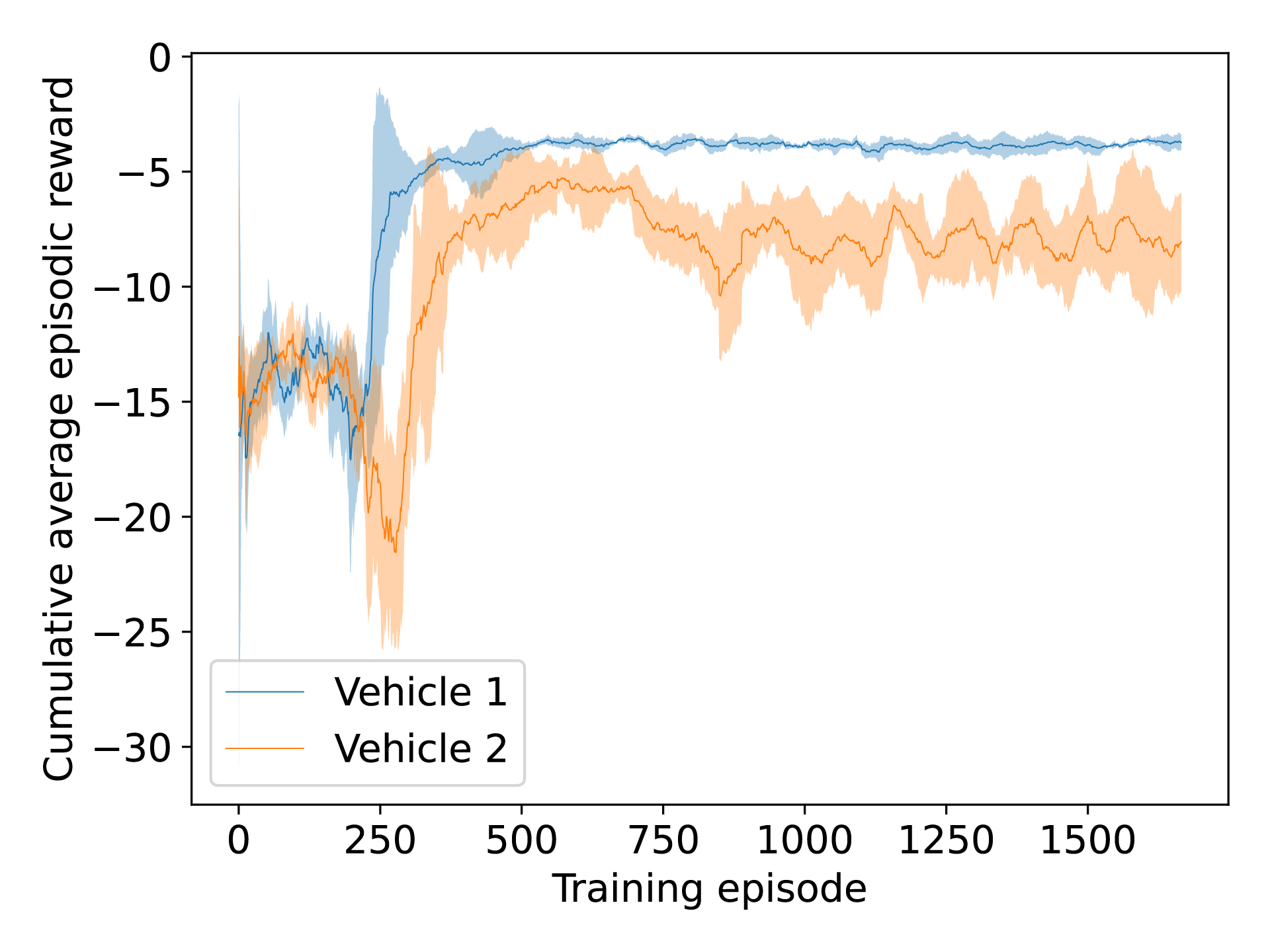}
       \caption{No-FRL: Platoon 1}\label{fig:avgep_pl1reward_nofrl}
    \end{subfigure} \hfill
    \begin{subfigure}{0.3\textwidth}

        \includegraphics[width=0.99\textwidth]{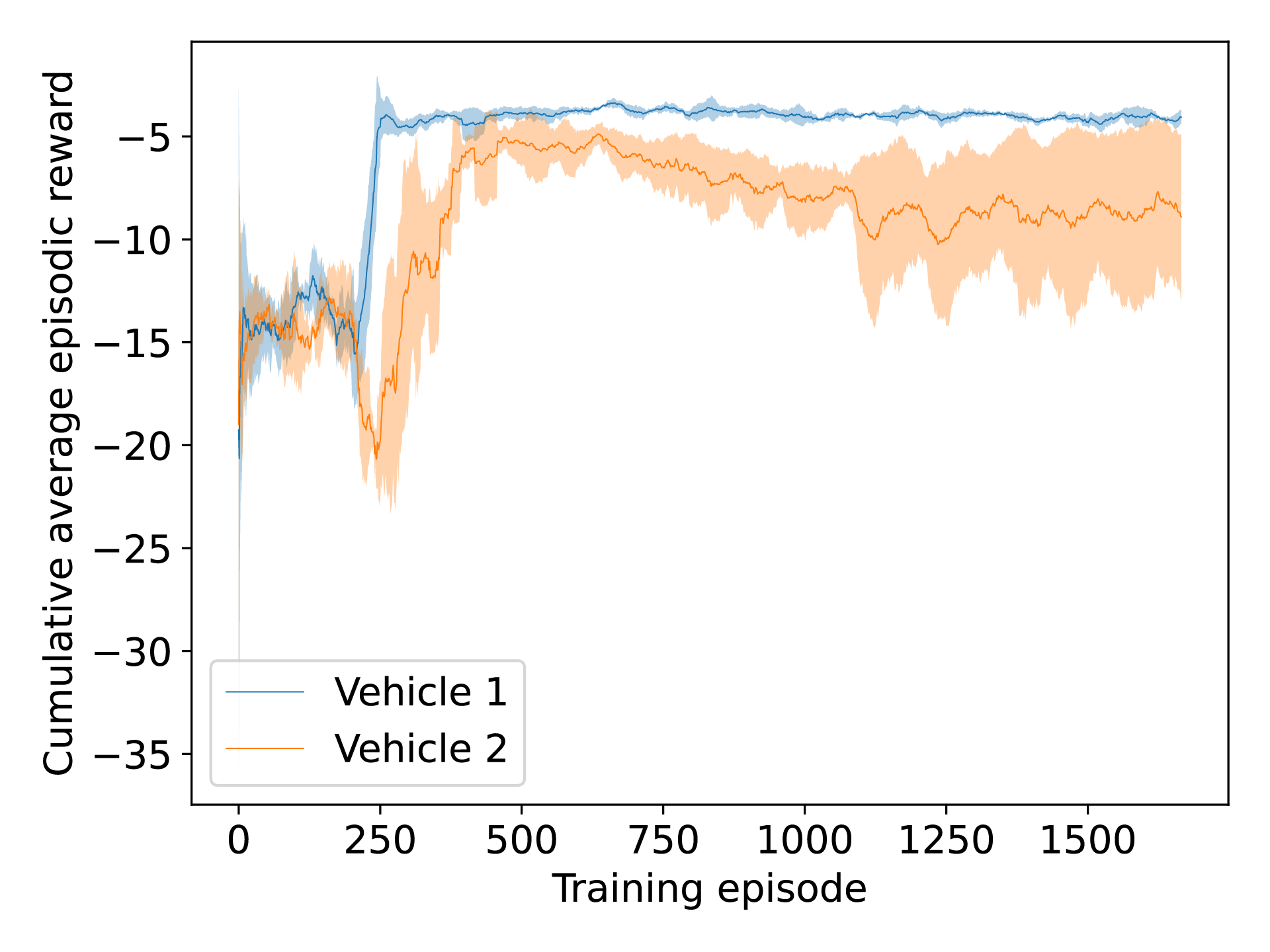}
       \caption{Inter-FRLGA: Platoon 1}\label{fig:avgep_pl1reward_grads}
    \end{subfigure}\hfill
    \begin{subfigure}{0.3\textwidth}

        \includegraphics[width=0.99\textwidth]{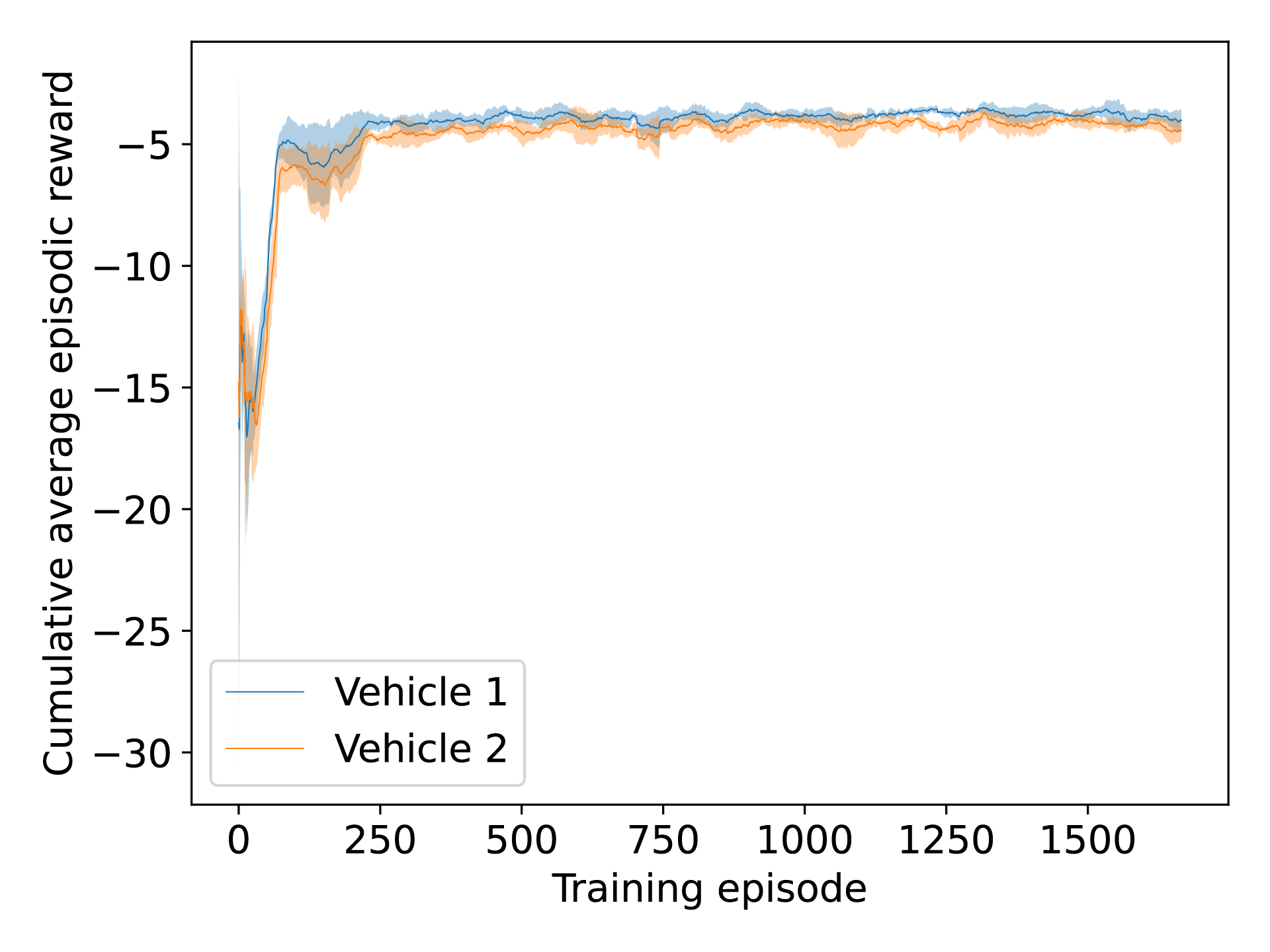}
       \caption{Inter-FRLWA: Platoon 1}\label{fig:avgep_pl1reward_weight}
    \end{subfigure}\hfill

    \begin{subfigure}{0.3\textwidth}

        \includegraphics[width=0.99\textwidth]{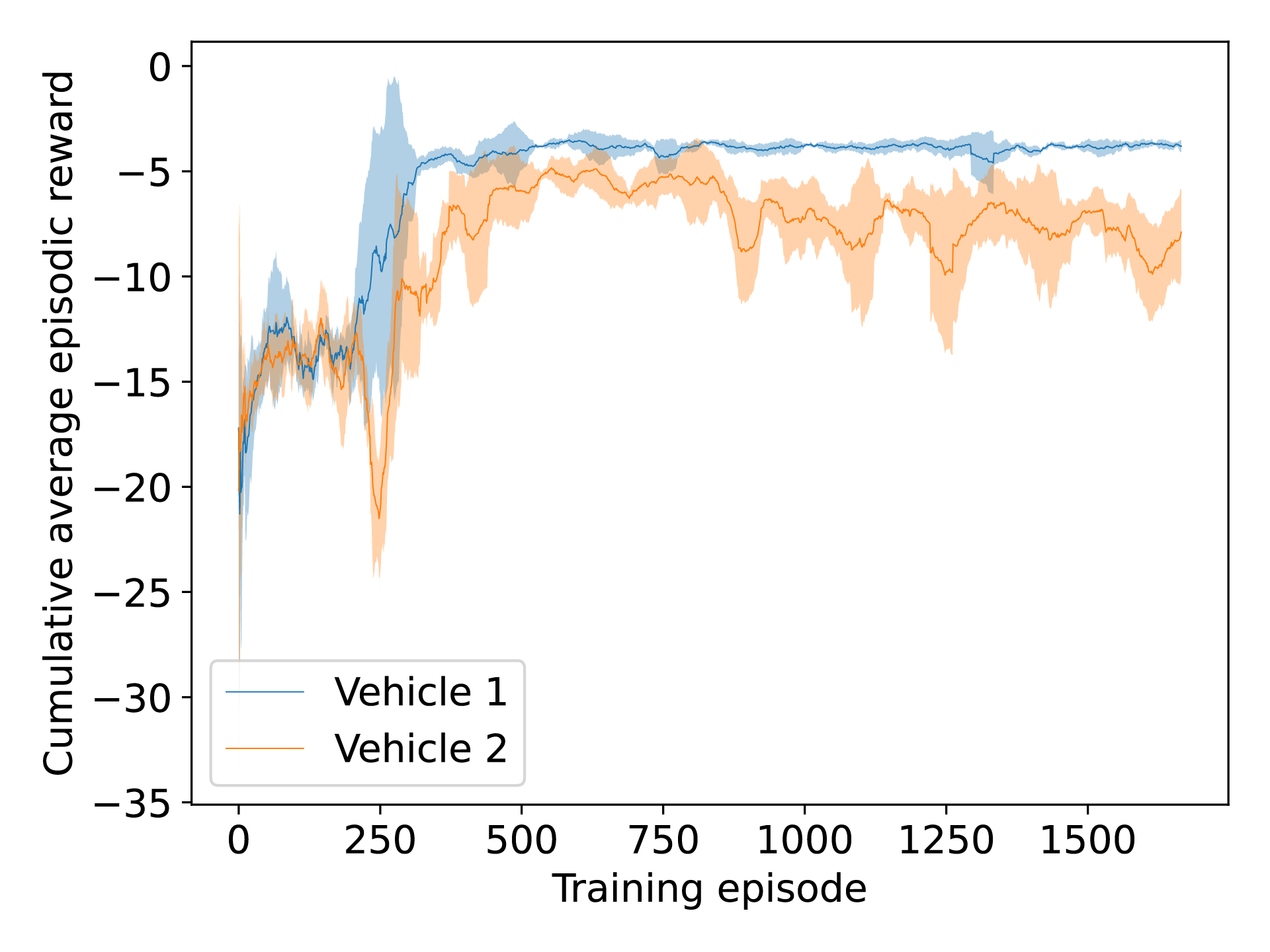}
       \caption{No-FRL: Platoon 2}\label{fig:avgep_pl2reward_nofrl}
    \end{subfigure}\hfill
    \begin{subfigure}{0.3\textwidth}

        \includegraphics[width=0.99\textwidth]{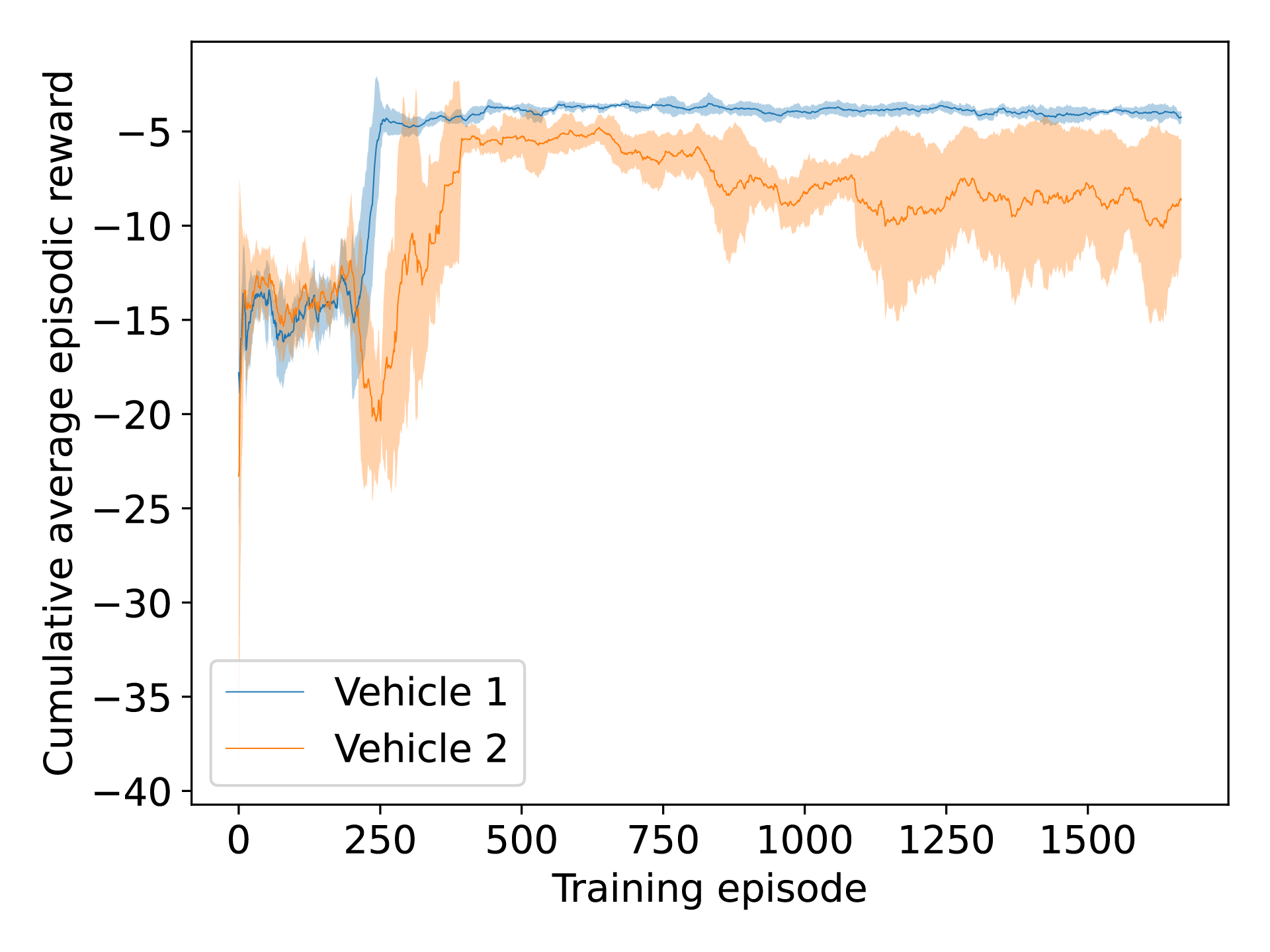}
       \caption{Inter-FRLGA: Platoon 2}\label{fig:avgep_pl2reward_grads}
    \end{subfigure}\hfill
    \begin{subfigure}{0.3\textwidth}
        \includegraphics[width=0.99\textwidth]{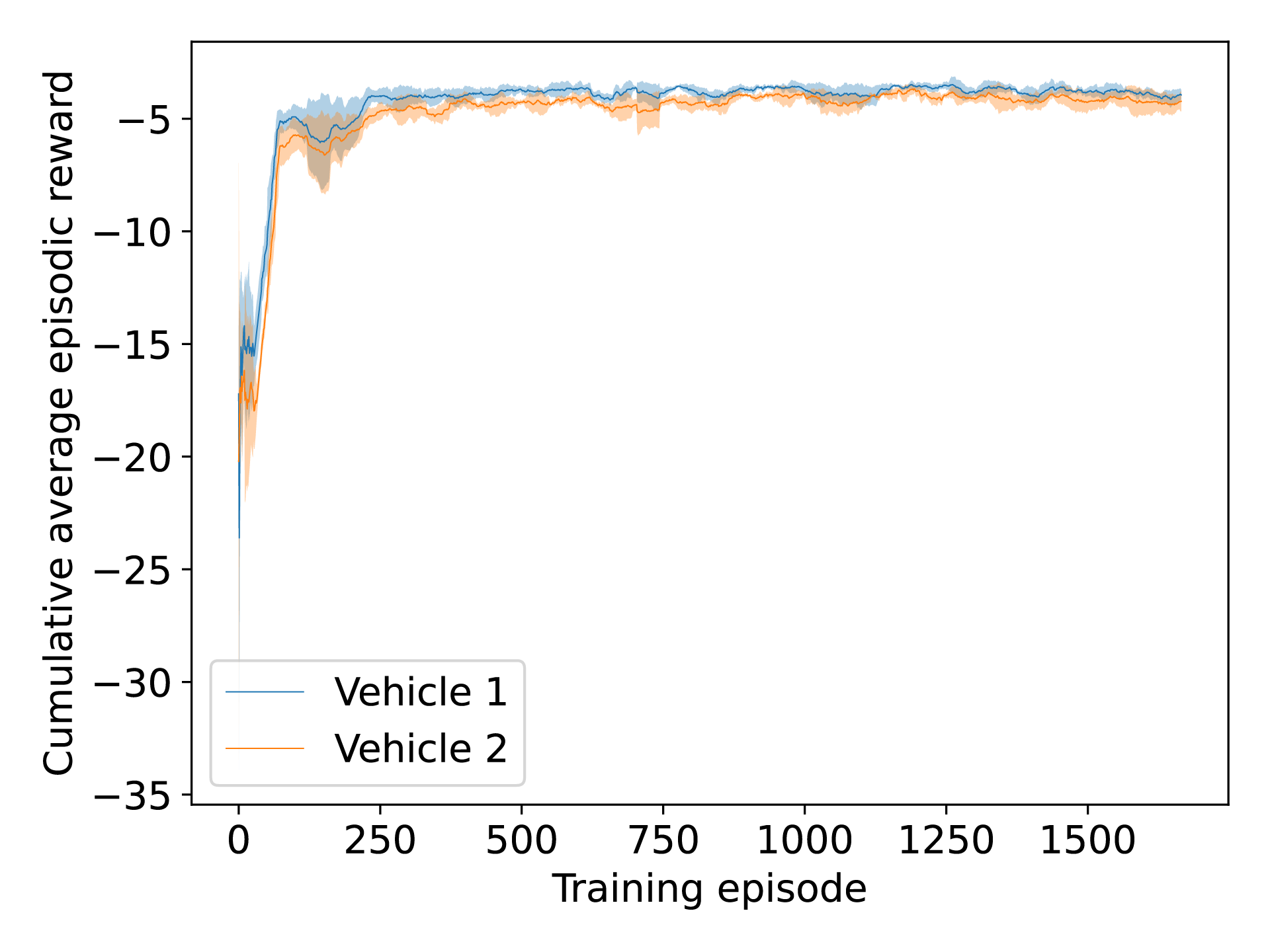}
       \caption{Inter-FRLWA: Platoon 2}\label{fig:avgep_pl2reward_weight}
    \end{subfigure}
\caption{Average performance across 4 random seeds for a 2 platoon 2 vehicle scenario trained without FRL (Figure \ref{fig:avgep_pl1reward_nofrl}, \ref{fig:avgep_pl2reward_nofrl}), with Inter-FRLGA (Figure \ref{fig:avgep_pl1reward_grads}, \ref{fig:avgep_pl2reward_grads}), and with Inter-FRLWA (Figure \ref{fig:avgep_pl1reward_weight}, \ref{fig:avgep_pl2reward_weight}). The shaded areas represent the standard deviation across the 4 seeds.}
\label{fig:inter_avep_rew}
\end{figure}

The cumulative reward is calculated over each training episode, and a moving average is computed over 40 episodes to generate Figure \ref{fig:avgep_pl1reward_nofrl}-\ref{fig:avgep_pl2reward_weight}.  It can be seen that the cumulative reward for Inter-FRLWA not only converges more rapidly than both no-FRL and Inter-FRLGA, but Inter-FRLWA also appears to have a more stable training session as indicated by the lower magnitude of the shaded area (the standard deviation across the four random seeds).

\subsubsection{Test results for one episode}
\begin{figure}[!t]
\centering
    \begin{subfigure}{0.45\textwidth}
        \centering
        \includegraphics[width=0.75\textwidth]{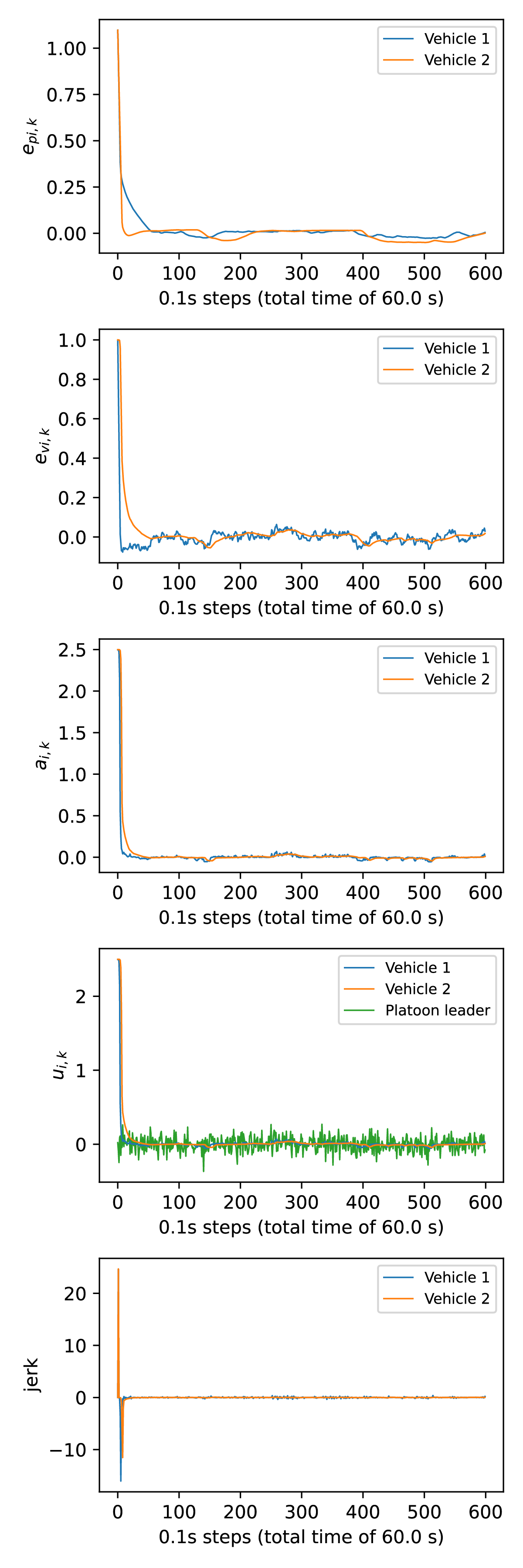}
       \caption{Platoon 1}\label{fig:simresINTERFRLWA_p1}
    \end{subfigure}\hspace{\interfrlRewSpace}
    \begin{subfigure}{0.45\textwidth}
        \centering
        \includegraphics[width=0.75\textwidth]{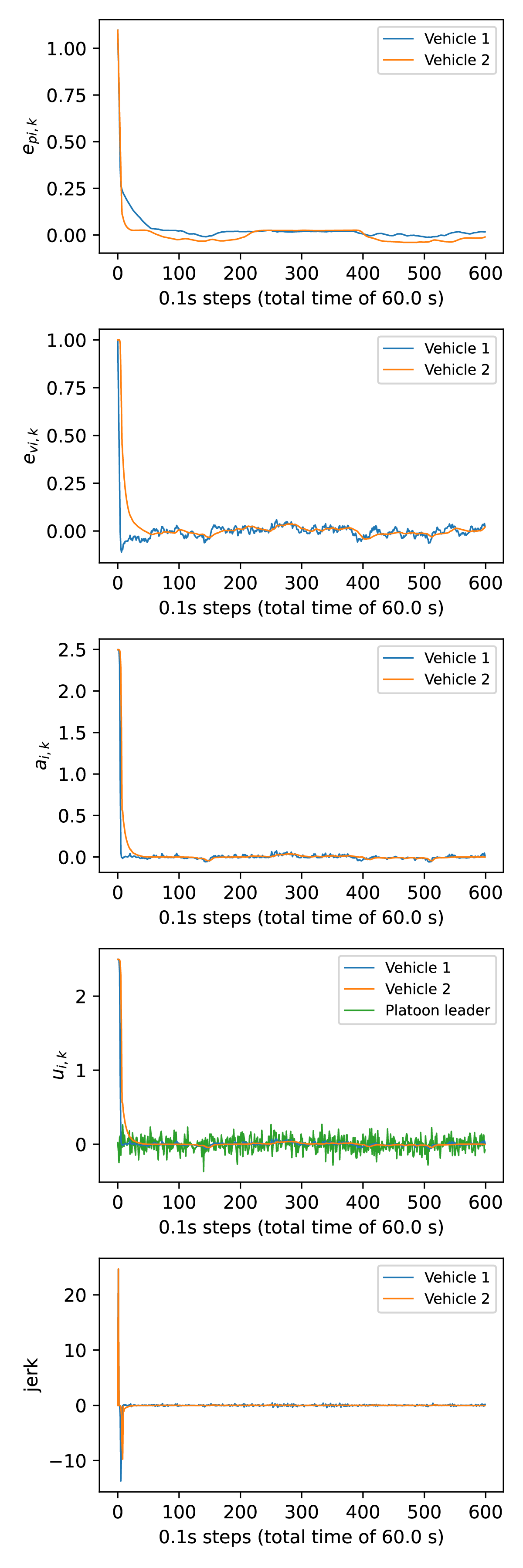}
       \caption{Platoon 2}\label{fig:simresINTERFRLWA_p2}
    \end{subfigure}
\label{fig:interfrlsim}
\caption{Results for a specific 60s test episode using the 2 vehicle 2 platoon environment trained using Inter-FRL with weight aggregation.}
\end{figure}

In Figure \ref{fig:simresINTERFRLWA_p1} and \ref{fig:simresINTERFRLWA_p2}, a simulation is performed over a single training episode plotting the jerk, along with the control input $u_{i,k}$, acceleration $a_{i,k}$, velocity error $e_{vi,k}$, and position error $e_{pi,k}$ for each platoon.  There are 2 platoons in the Inter-FRL scenario, and a simulation is provided for each platoon. The simulation environment is subject to initial conditions of ($e_{pi} = 1.0\:m$, $e_{vi}=1.0\:m/s$, $a_i = 0.03\:m/s^2$).  It can be seen that each DDPG agent for both vehicles within both platoons quickly responds to the platoon leader's control input $u_{i,k}$ to bring the position error, velocity error and acceleration error to 0.  In addition, each DDPG agent closely approximates the Gaussian random input of the platoon leader, eliminating noise in the response to maintain smooth tracking across the episode. Finally, each DDPG agent in the platoon also minimizes the jerk effectively. These results are indicative of both a good design of the reward function \eqref{eqn:rewardB}, and also a suitable selection of parameters $a, b, c$ and $d$ in \eqref{eqn:rewardB}.

\subsection{Intra-FRL}
In order to evaluate the effectiveness of Intra-FRL relative to the base AV platooning scenario, 4 experiments are conducted without Intra-FRL (no-FRL), and 8 with.  For each of the conducted experiments, 1 platoon with 2 vehicles is trained using 4 random seeds.  A single platoon is required for studying Intra-FRL as parameters are shared amongst vehicles within the platoon (no sharing is performed from vehicle's in one platoon to another).  Once training across the four seeds is completed, the cumulative reward for a single evaluation episode is evaluated.  Similar to the experiments using Inter-FRL, two aggregation methods are examined.  First, the gradients of each model are averaged during training, and second, the model weights are averaged.  The platoon trains and shares the aggregated parameters (gradients or weights) from vehicle to vehicle such that data is averaged and updated amongst vehicles within the same platoon. The federated server is responsible for performing the averaging, and each vehicle performs a training episode with the averaged parameters in addition to their local training episodes in accordance with the FRL update delay and FRL cutoff ratio (see Table \ref{tab:frlhyps}).  Note that here-after Intra-FRL with gradient aggregation is denoted Intra-FRLGA, and Intra-FRL with weight aggregation is denoted Intra-FRLWA.

\subsubsection{Performance across 4 random seeds}
The performance for the platoon is calculated by averaging the cumulative reward generated by the simulation for each of the 4 random seeds and is summarized in Table \ref{tab:intrafrl-summary}. The results in Table \ref{tab:intrafrl-summary} summarize the performance for no-FRL, Intra-FRLGA, and lastly Intra-FRLWA.  It is observed that Intra-FRLWA performs most favourably, followed by no-FRL and lastly Intra-FRLGA.

\begin{table*}[!t]
  \centering
  \caption{Performance after training across 4 random seeds. Each simulation result contains 600 time steps.}
    \begin{tabular}{lrrrrrr} \toprule
    \textbf{Training method} & \multicolumn{1}{l}{\textbf{Seed 1}} & \multicolumn{1}{l}{\textbf{Seed 2}} & \multicolumn{1}{l}{\textbf{Seed 3}} & \multicolumn{1}{l}{\textbf{Seed 4}} & \multicolumn{1}{l}{\textbf{Average system reward}} & \multicolumn{1}{l}{\textbf{Standard deviation}} \\ \midrule
    No-FRL & -3.84 & -3.40 & -3.29 & -3.21 & -3.44 & 0.24 \\
    Intra-FRLGA & -2.85 & -8.05 & -4.23 & -2.99 & -4.53 & 2.10 \\
    Intra-FRLWA & -2.56 & -2.60 & -2.68 & -2.75 & -2.65 & 0.07 \\ \bottomrule
    \end{tabular}%
  \label{tab:intrafrl-summary}%
\end{table*}%

 \subsubsection{Convergence properties}

\begin{figure}[!t]
\centering
    \begin{subfigure}{\interfrlRewWidth\textwidth}
        \raggedleft
        \includegraphics[width=\textwidth]{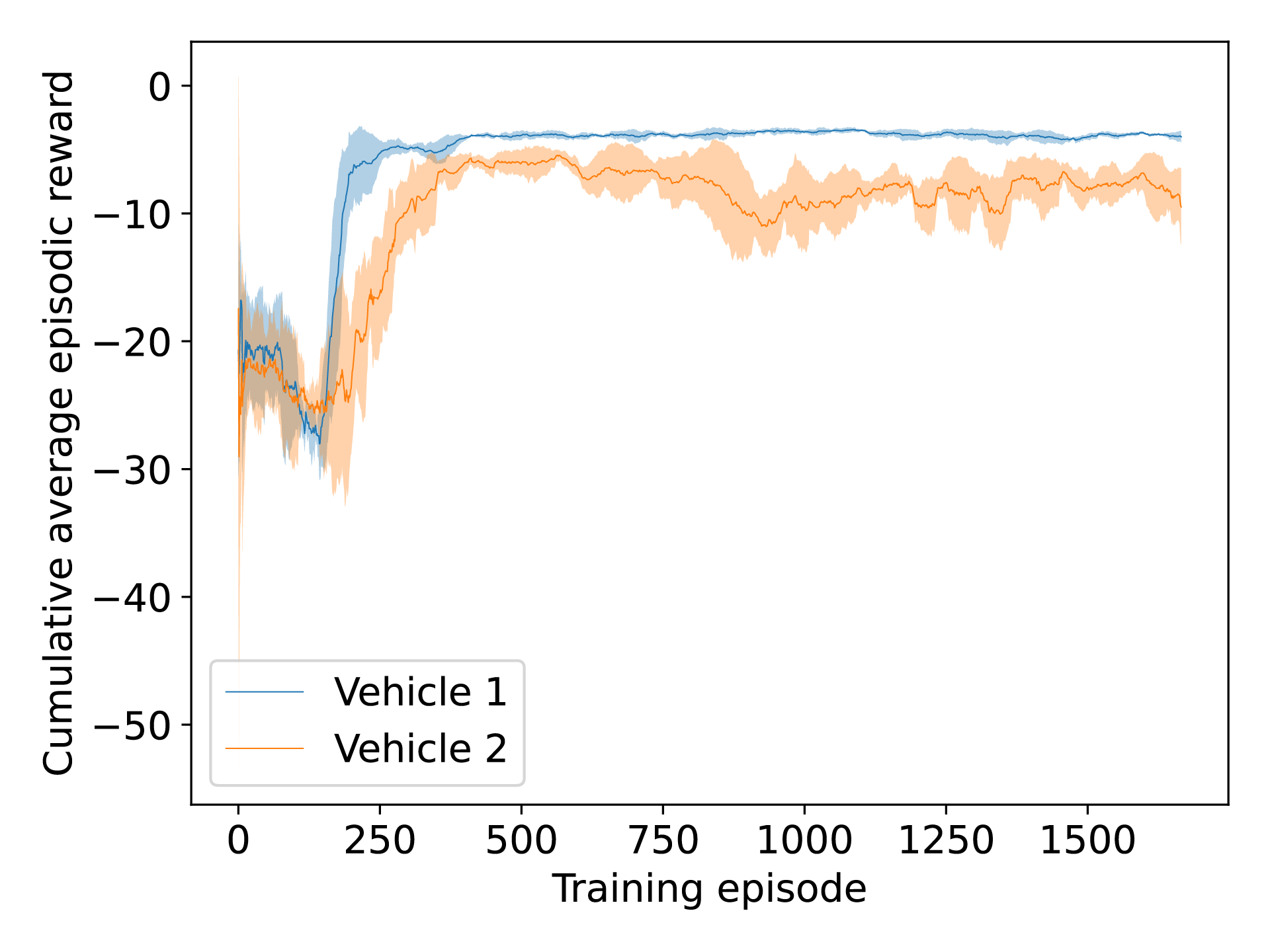}
       \caption{No-FRL}\label{fig:intra_avgep_pl1reward_nofrl}
    \end{subfigure}\hspace{\interfrlRewSpace}
    \begin{subfigure}{\interfrlRewWidth\textwidth}
        \raggedleft
        \includegraphics[width=\textwidth]{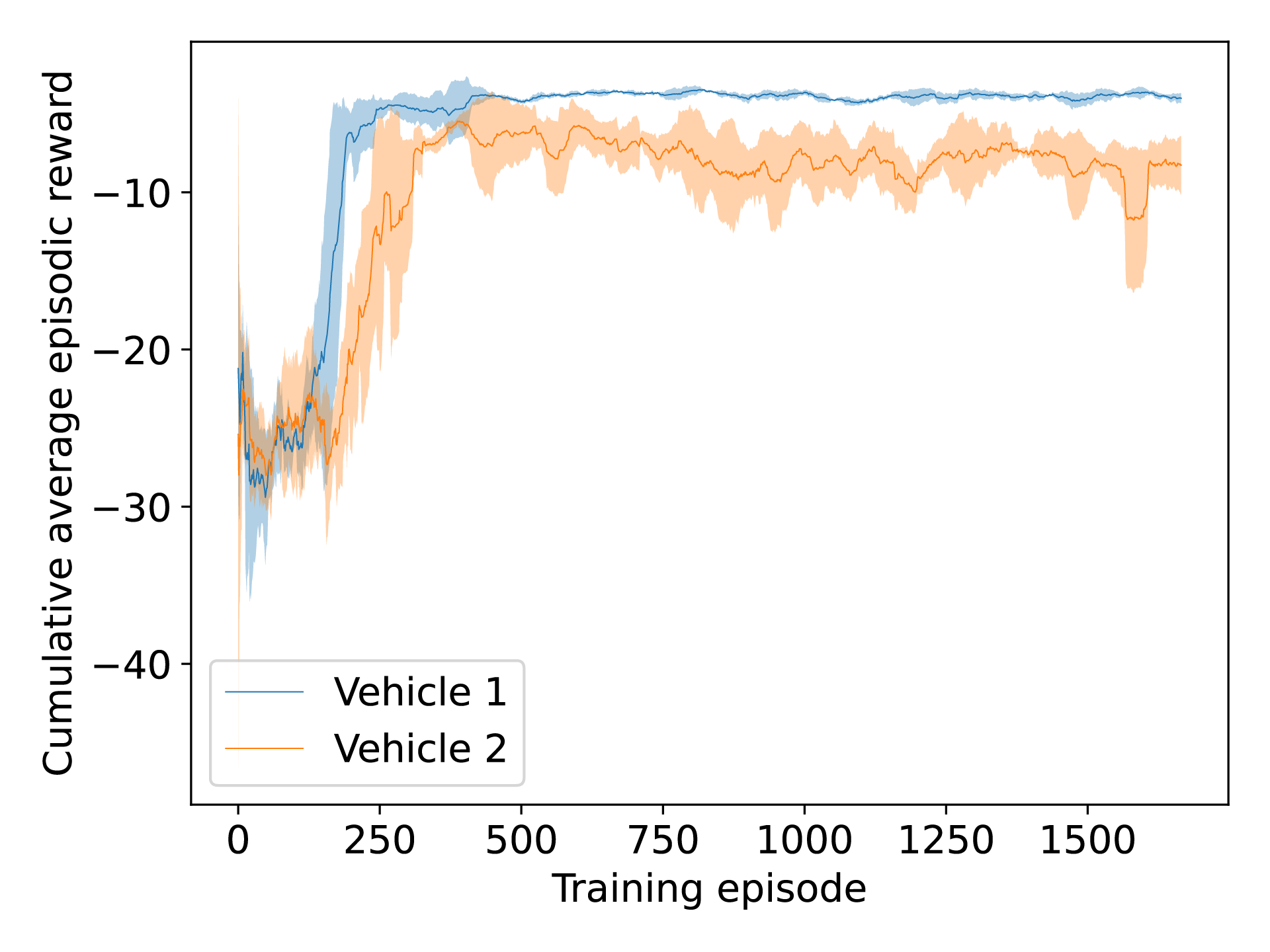}
       \caption{Intra-FRLGA}\label{fig:intra_avgep_pl1reward_grads}
    \end{subfigure}\hspace{\interfrlRewSpace}
        \begin{subfigure}{\interfrlRewWidth\textwidth}
        \raggedleft
        \includegraphics[width=\textwidth]{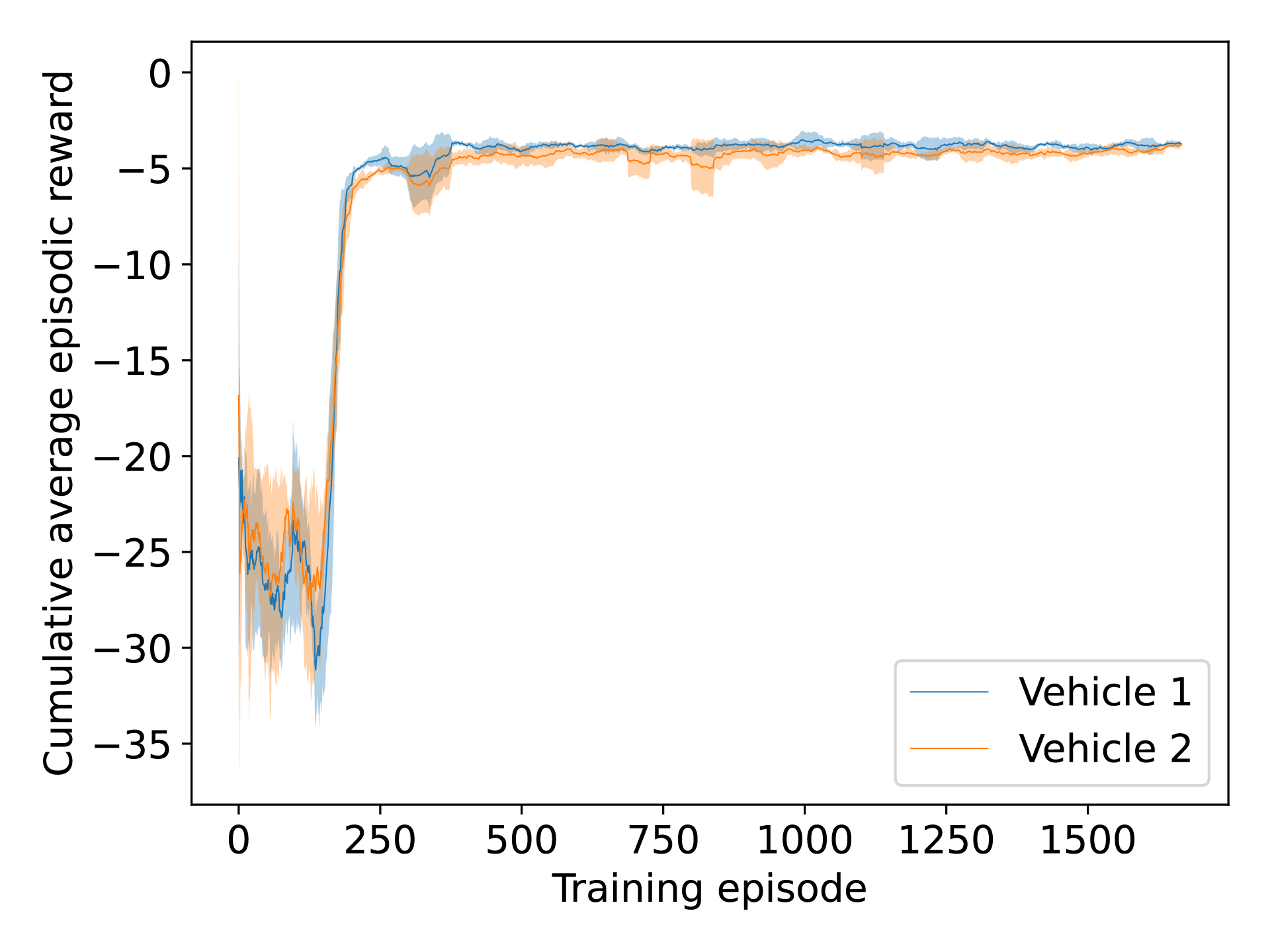}
       \caption{Intra-FRLWA}\label{fig:intra_avgep_pl1reward_weight}
    \end{subfigure}
\caption{Average performance across 4 random seeds for 1 platoon 2 vehicle scenario trained without FRL (Figure \ref{fig:avgep_pl1reward_nofrl}), Intra-FRLGA (Figure \ref{fig:avgep_pl1reward_grads}), and with Intra-FRLWA (Figure \ref{fig:avgep_pl1reward_weight}). The shaded areas represent the standard deviation across the 4 seeds.}
\label{fig:intra-convergence}
\end{figure}

The cumulative reward is calculated over each training episode, and a moving average is computed over 40 episodes to generate Figure \ref{fig:intra-convergence}.  Similar to the Inter-FRL experiments, Intra-FRLWA shows the most favourable training results.  In addition, the rate of convergence increases with Intra-FRLWA over no-FRL and Intra-FRLGA.  Lastly, the stability during training is also shown to be improved as the standard deviation across the four random seeds is much smaller than the other two cases (as evident in the shaded regions of Figure \ref{fig:intra-convergence}).

\subsubsection{Test results for one episode}
\begin{figure}[!t]
    \centering
    \includegraphics[width=0.35\textwidth]{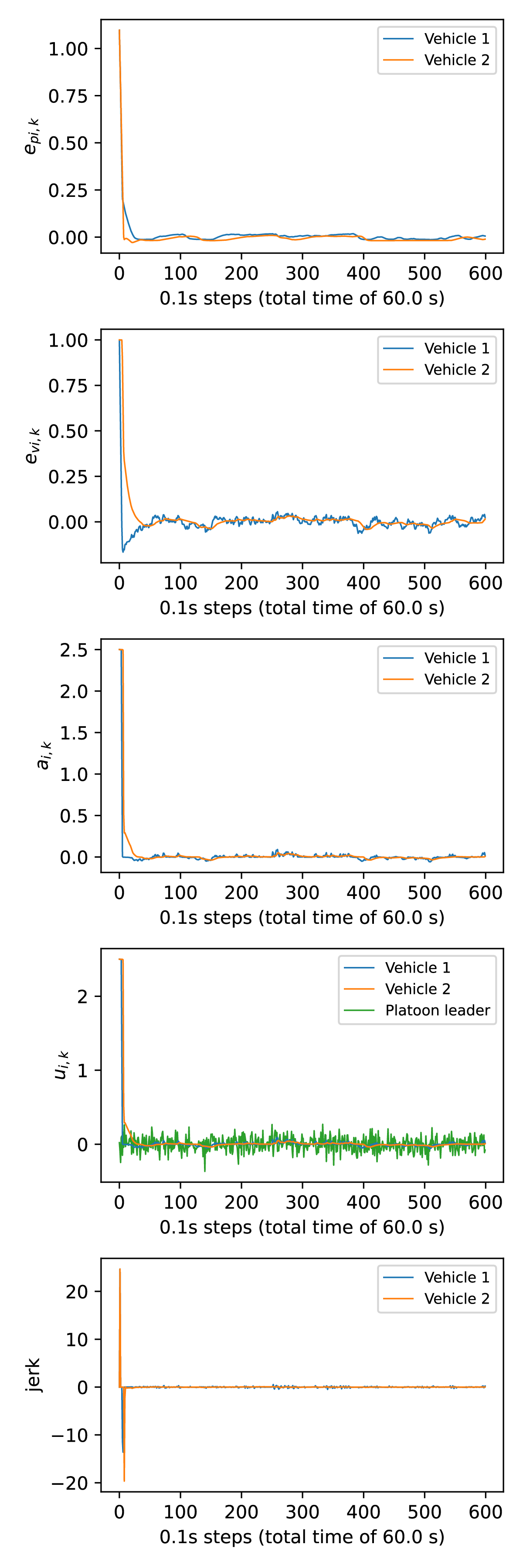}
    \caption{Results for a specific 60s test episode using the 2 vehicle 1 platoon environment trained using Intra-FRLWA.}
    \label{fig:intraFRL-simresult}
\end{figure}

A single simulation is performed on an episode plotting the jerk, along with the control input $u_{i,k}$, acceleration $a_{i,k}$, velocity error $e_{vi,k}$, and position error $e_{pi,k}$.  Figure \ref{fig:intraFRL-simresult} shows the precise control of Intra-FRLWA on the environment.  The environment is initialized to the same conditions as that of the Inter-FRLWA scenario ($e_{pi} = 1.0 m$, $e_{vi}=1.0 m/s$, $a_i = 0.03 m/s^2$), and each DDPG agent in the platoon quickly and precisely tracks the Gaussian random input $u_{i,k}$ from the leader while minimizing position error, velocity error, acceleration , and jerk.  Much like the Inter-FRLWA scenario, it is observed that a strong optimization of the reward function (Equation 10) has occurred.  This is an indication of a good design of the reward function in addition to a good balance of parameters $a,b,c$ and $d$ in the reward function.

\subsection{Comparison between inter and intra-FRL}
The results for both Inter-FRL and Intra-FRL are summarized in Table \ref{tab:inter_vs_intra} below.

\begin{table*}[!t]
    \centering
    \caption{Performance after training across 4 random seeds for both Inter and Intra FRL. Each simulation result contains 600 time steps.}
    \begin{tabular}{lrrlrrr} \toprule
    \textbf{Training Method} & \multicolumn{1}{l}{\textbf{Seed 1}} & \multicolumn{1}{l}{\textbf{Seed 2}} & \multicolumn{1}{l}{\textbf{Seed 3}} & \multicolumn{1}{l}{\textbf{Seed 4}} & \multicolumn{1}{l}{\textbf{Average system reward}} & \multicolumn{1}{l}{\textbf{Standard deviation}} \\ \midrule
        Inter-FRLGA & -2.79 & -2.81 & -3.05 & -2.76 & -2.85 & 0.11 \\
        Inter-FRLWA & -2.64 & -2.88 & -2.92 & -2.93 & -2.84 & 0.12 \\
        Intra-FRLGA & -2.85 & -8.05 & -4.23 & -2.99 & -4.53 & 2.10 \\
        Intra-FRLWA & -2.56 & -2.60 & -2.68 & -2.75 & -2.65 & 0.07 \\ \bottomrule
    \end{tabular}
    \label{tab:inter_vs_intra}
\end{table*}

It is clear that using weight aggregation in both Inter-FRL and Intra-FRL is favourable to gradient aggregation.  In addition, Intra-FRLWA provides the overall best result.  Intra-FRL likely converges to the best model due to conditions each agent experiences during training.  For Inter-FRL, the environment is independent and identically distributed. For Intra-FRL, each follower's training depends on the policy of the preceding vehicle.  For the 2 vehicle scenario studied, vehicle 1 will converge prior to vehicle 2 as vehicle 1 learns based on the stochastic random input generated by the platoon leader. As vehicle 1 is training, vehicle 2 trains based off the policy of vehicle 1. As previously stated, Inter-FRL shares parameters amongst vehicles in the same index across platoons, where-as Intra-FRL provides the advantage of sharing parameters from preceding vehicles to following vehicles. Our implementation of Intra-FRL includes a directional parameter averaging.  For example, vehicle 1 does not train with averaged parameters from the followers, but vehicle 2 has the advantage of including vehicle 1's model in its averaging.  This directional averaging provides an advantage to vehicle 2, as evidenced by the increased performance in Table \ref{tab:inter_vs_intra}.

\subsection{Intra-FRL with variant number of vehicles}
An additional factor to consider when evaluating FRL in relation to the no-FRL base scenario is how FRL performs with increasing agents relative to no-FRL.  In this section, 12 experiments are conducted with no-FRL, and 12 with Intra-FRLWA. Each set of 12 experiments for no-FRL and Intra-FRLWA are broken up by number of vehicles and random seed.  The random seed is selected to be a value between 1 and 4, inclusive. In addition, the platoons under study contain either 3, 4, or 5 vehicles. Once training has been completed for all experiments, the cumulative reward for each experiment is evaluated using a single simulation episode in which the seed is kept constant.  Intra-FRLWA is used as the FRL training strategy since Intra-FRLWA was identified to be the highest performing FRL strategy in the previous section.

\subsubsection{Performance with varying number of vehicles}
The performance for each experiment is calculated by taking the average cumulative episodic reward across each vehicle in the platoon at the end of the simulation episode.  Table \ref{tab:varypl} presents the results for no-FRL and Intra-FRLWA for platoons with 3, 4, and 5 follower vehicles. Table \ref{tab:varypl} shows that Intra-FRLWA provides favourable performance in all platoon lengths. A notable example of Intra-FRLWA's success is highlighted when considering the poor performance of the 4 vehicle platoon trained with no-FRL using seed 1. The Intra-FRLWA training strategy was able to overcome the performance challenges, correcting the poor performance entirely.

\begin{table}[!t]
  \centering
  \caption{Performance after training across 4 random seeds with varying platoon lengths. Each simulation result contains 600 time steps.}
    \begin{tabular}{lrrrrrrr} \toprule
    \textbf{Training Method} & \multicolumn{1}{l}{\textbf{No. Vehicles}} & \multicolumn{1}{l}{\textbf{Seed 1}} & \multicolumn{1}{l}{\textbf{Seed 2}} & \multicolumn{1}{l}{\textbf{Seed 3}} & \multicolumn{1}{l}{\textbf{Seed 4}} & \multicolumn{1}{l}{\textbf{Avg. System Reward}} & \multicolumn{1}{l}{\textbf{Std. Dev.}} \\ \midrule
    No-FRL & 3     & -3.64 & -3.28 & -3.76 & -3.52 & -3.55 & 0.20 \\
    No-FRL & 4     & -123.58 & -4.59 & -7.39 & -4.51 & -35.02 & 59.06 \\
    No-FRL & 5     & -4.90 & -5.94 & -6.76 & -6.11 & -5.93 & 0.77 \\
    Intra-FRLWA & 3     & -3.44 & -3.16 & -3.43 & -4.14 & -3.54 & 0.42 \\
    Intra-FRLWA & 4     & -3.67 & -3.56 & -4.10 & -3.60 & -3.73 & 0.25 \\
    Intra-FRLWA & 5     & -3.92 & -4.11 & -4.33 & -3.97 & -4.08 & 0.18 \\ \bottomrule
    \end{tabular}%
  \label{tab:varypl}%
\end{table}%

\subsubsection{Convergence properties}
\begin{figure}[!t]
\centering
    \begin{subfigure}{\interfrlRewWidth\textwidth}
        \raggedleft
        \includegraphics[width=0.99\textwidth]{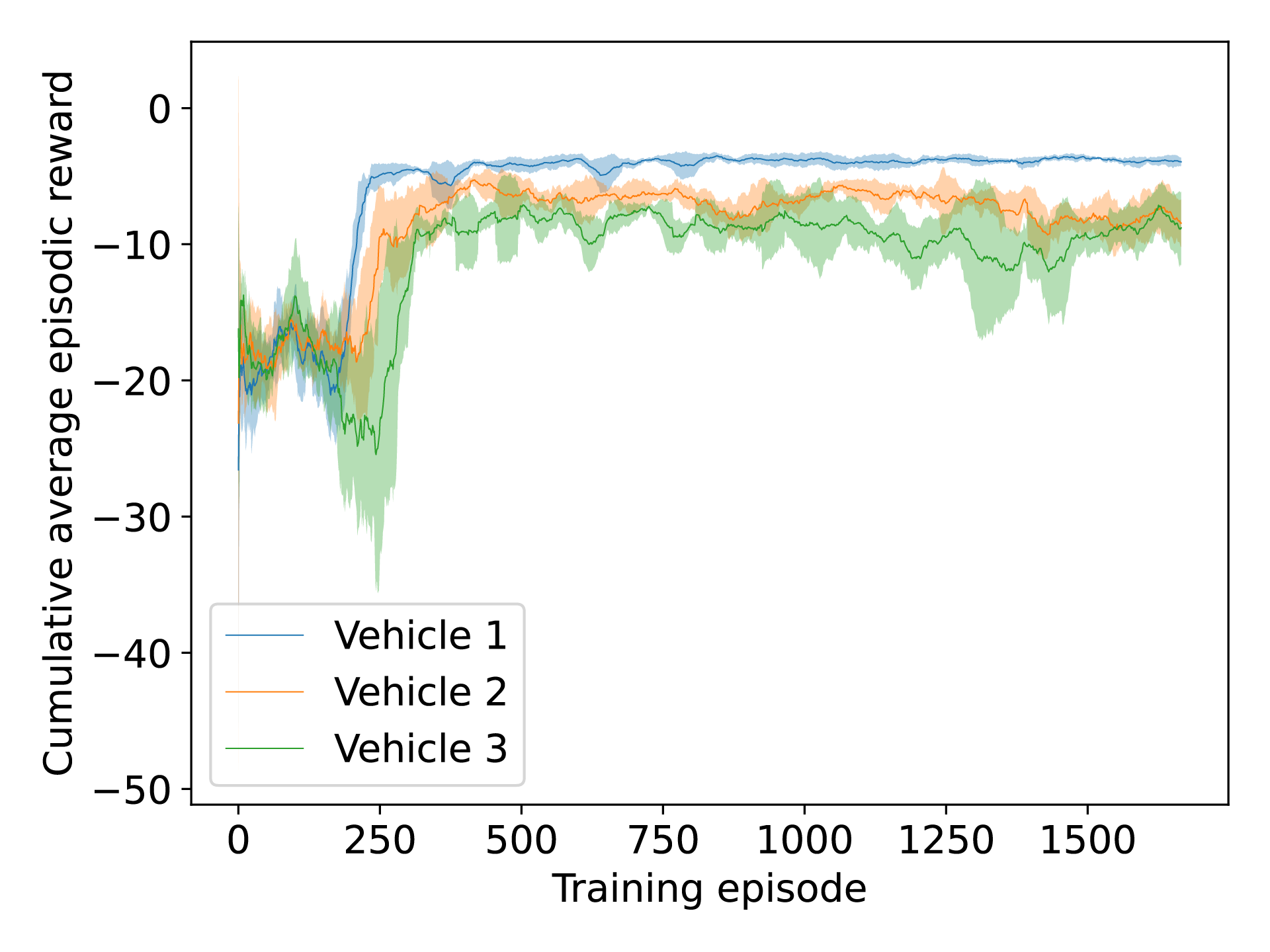}
      \caption{No-FRL: 3 Vehicles}\label{fig:reward_curve_p1_3v_nofrl}
    \end{subfigure}\hspace{\interfrlRewSpace}
    \begin{subfigure}{\interfrlRewWidth\textwidth}
        \raggedleft
        \includegraphics[width=0.99\textwidth]{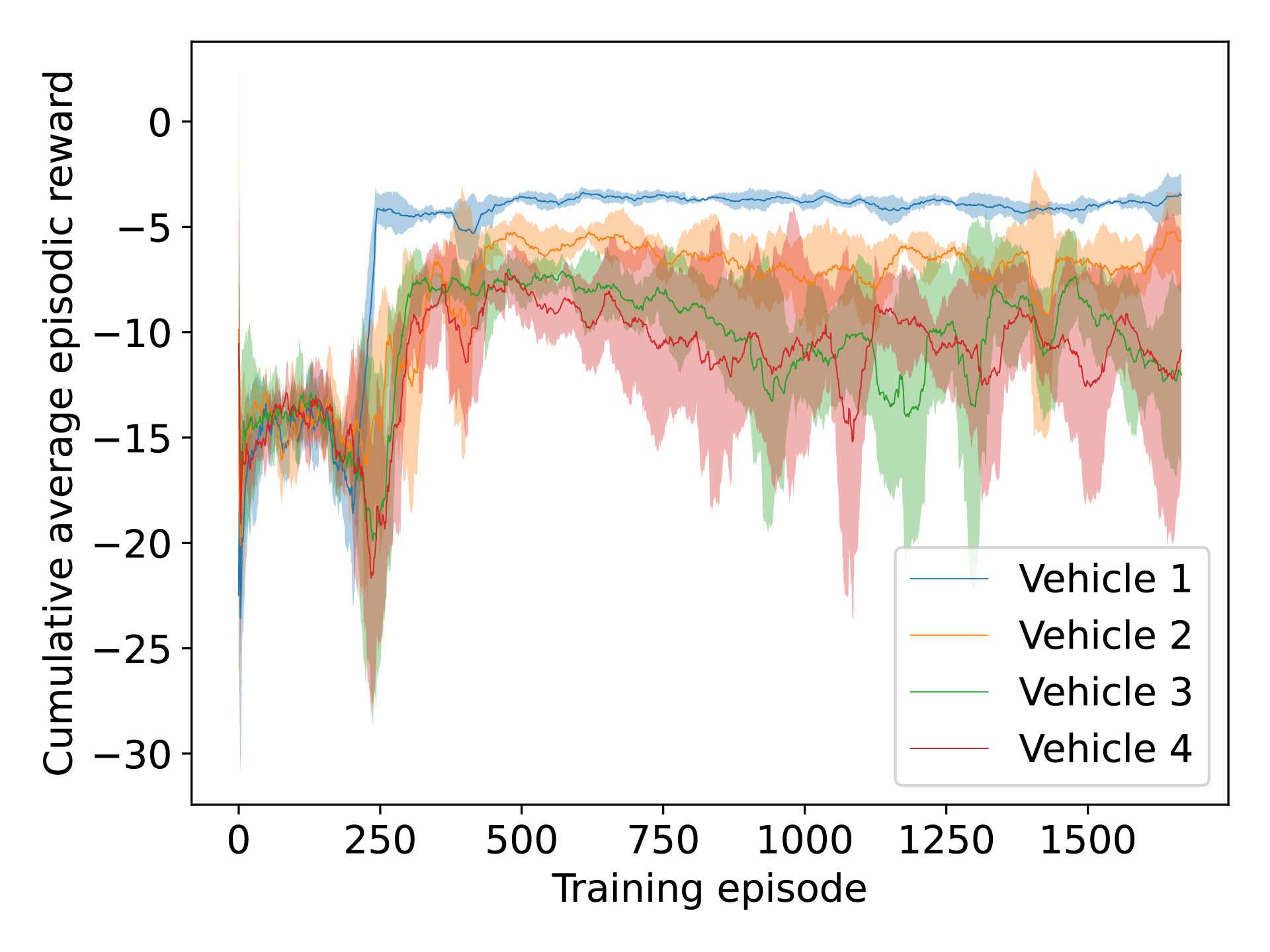}
      \caption{No-FRL: 4 Vehicles}\label{fig:reward_curve_p1_4v_nofrl}
    \end{subfigure}
    \begin{subfigure}{\interfrlRewWidth\textwidth}
        \raggedleft
        \includegraphics[width=0.99\textwidth]{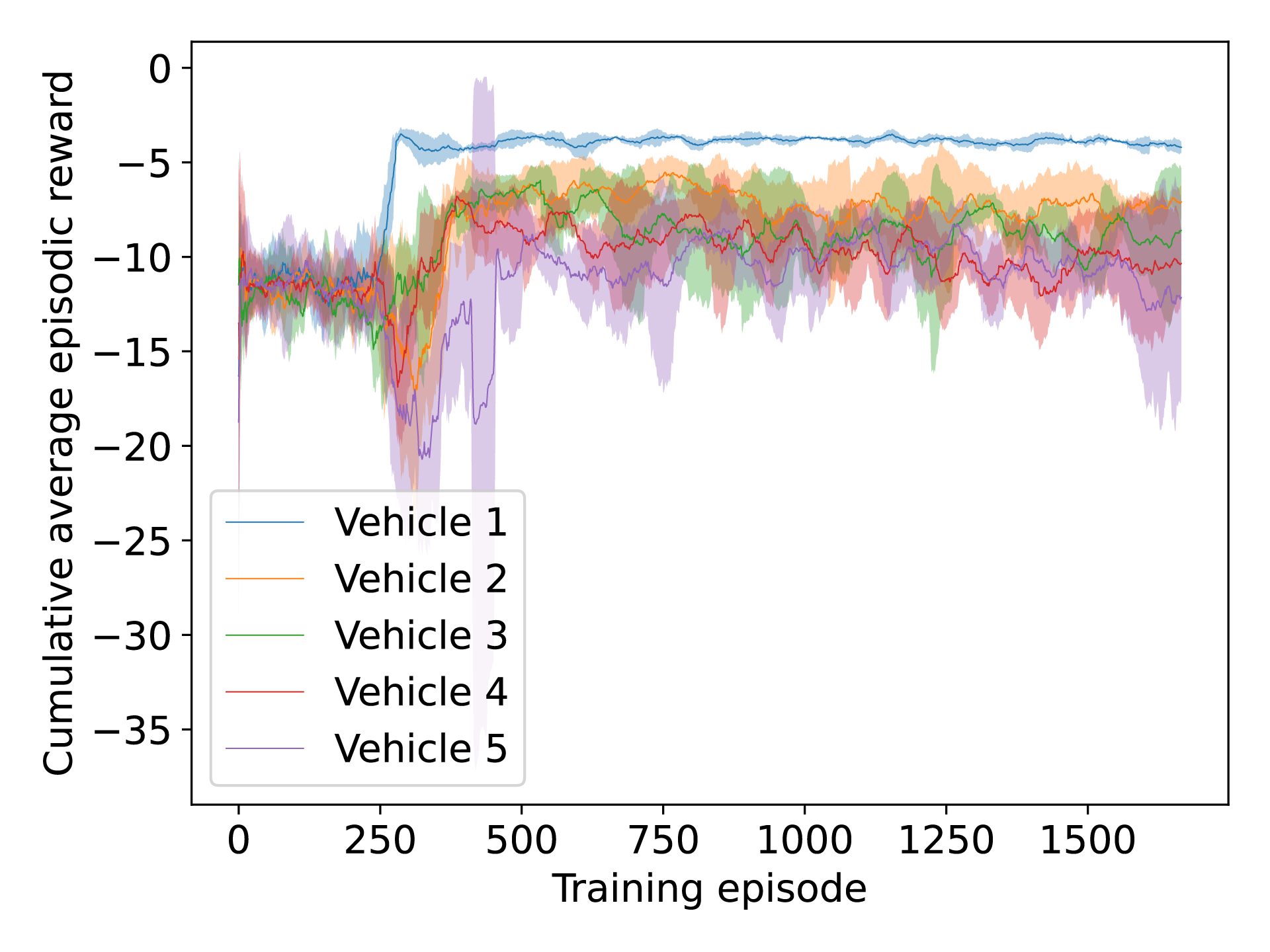}
      \caption{No-FRL: 5 Vehicles}\label{fig:reward_curve_p1_5v_nofrl}
    \end{subfigure}\hspace{\interfrlRewSpace}
    \begin{subfigure}{\interfrlRewWidth\textwidth}
        \raggedleft
        \includegraphics[width=0.99\textwidth]{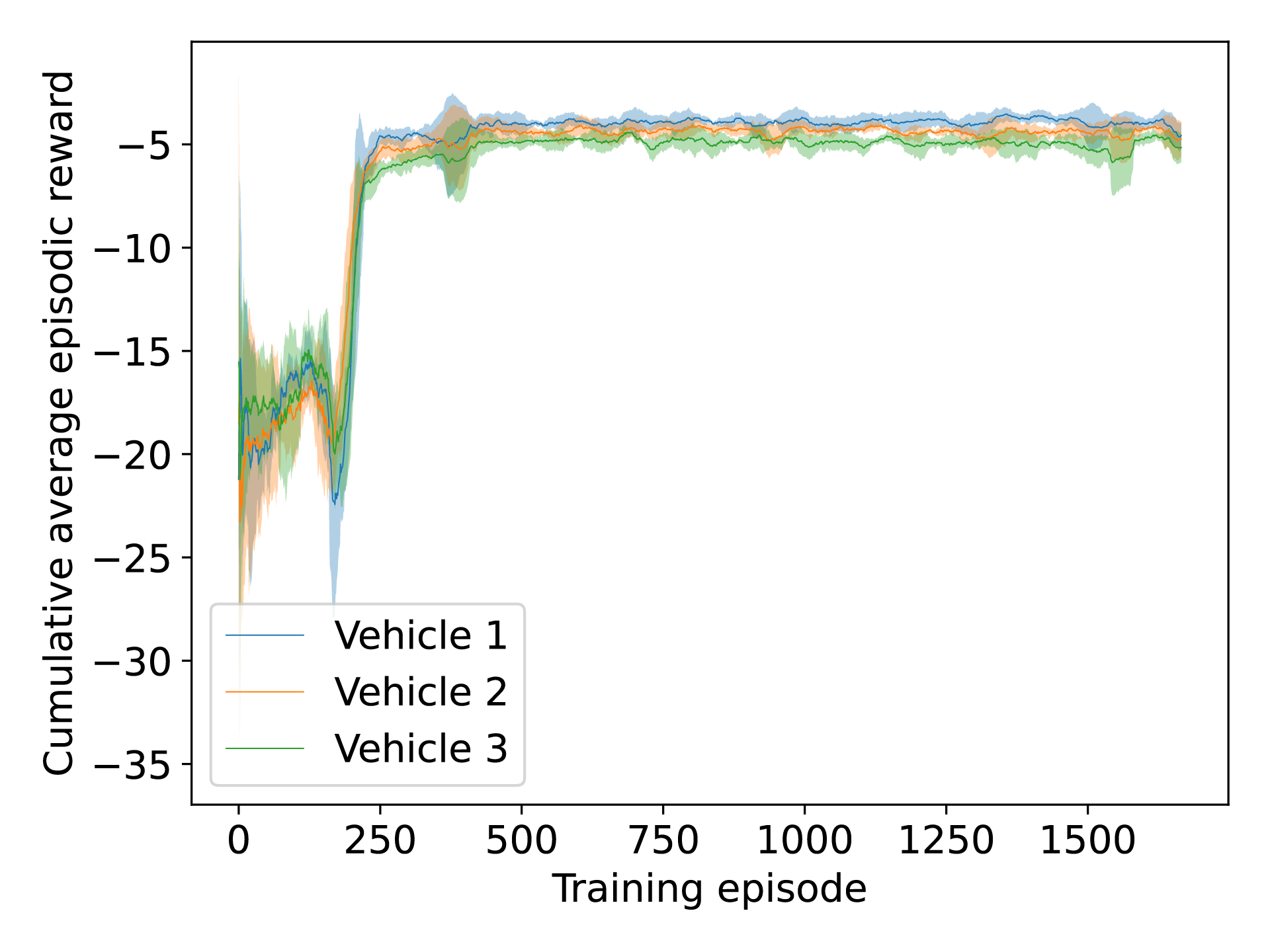}
      \caption{Intra-FRLWA: 3 Vehicles}\label{fig:reward_curve_p1_3v_intrafrlwa}
    \end{subfigure}\hspace{\interfrlRewSpace}
    \begin{subfigure}{\interfrlRewWidth\textwidth}
        \raggedleft
        \includegraphics[width=0.99\textwidth]{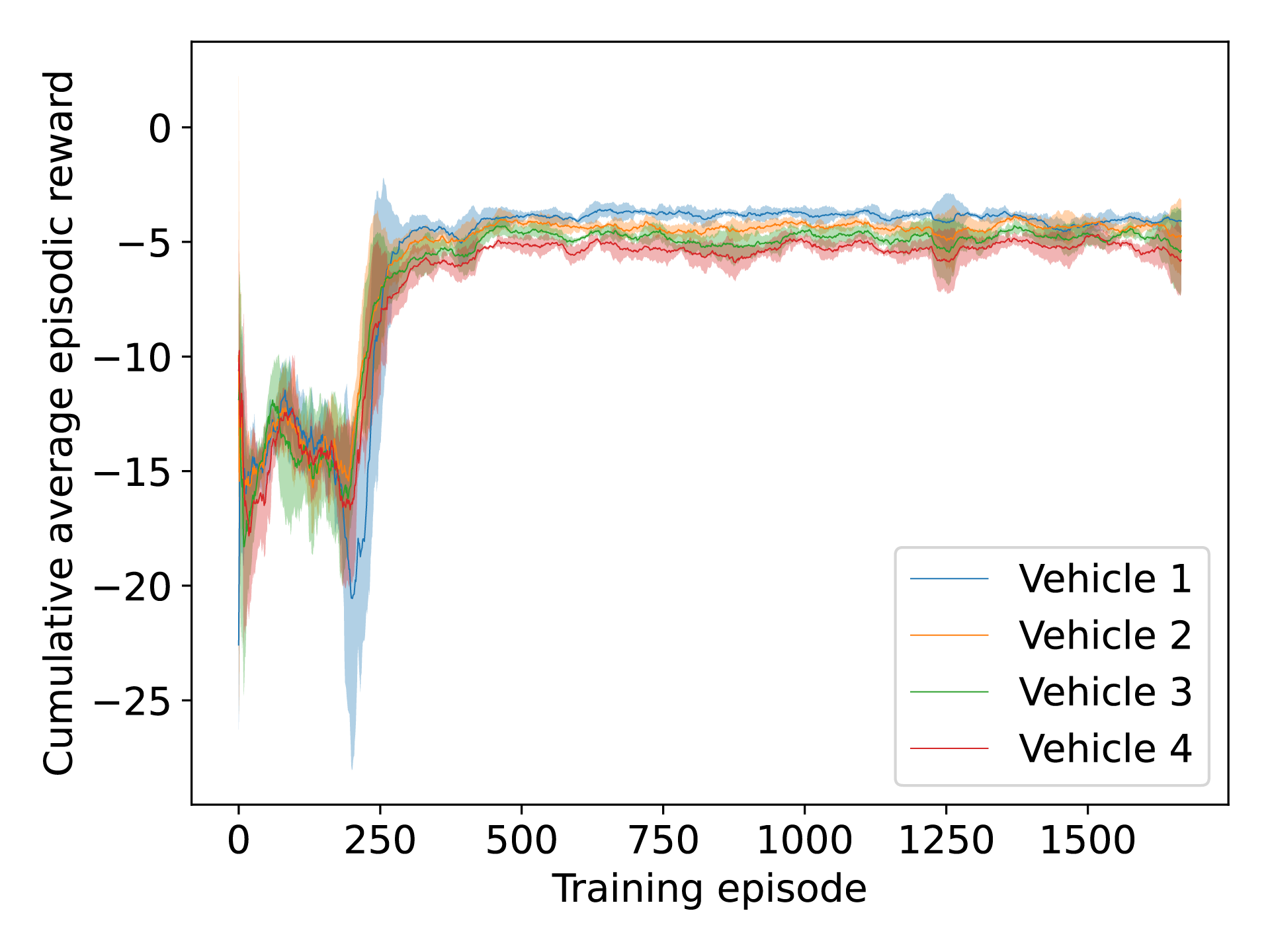}
      \caption{Intra-FRLWA: 4 Vehicles}\label{fig:reward_curve_p1_4v_intrafrlwa}
    \end{subfigure}\hspace{\interfrlRewSpace}
    \begin{subfigure}{\interfrlRewWidth\textwidth}
        \raggedleft
        \includegraphics[width=0.99\textwidth]{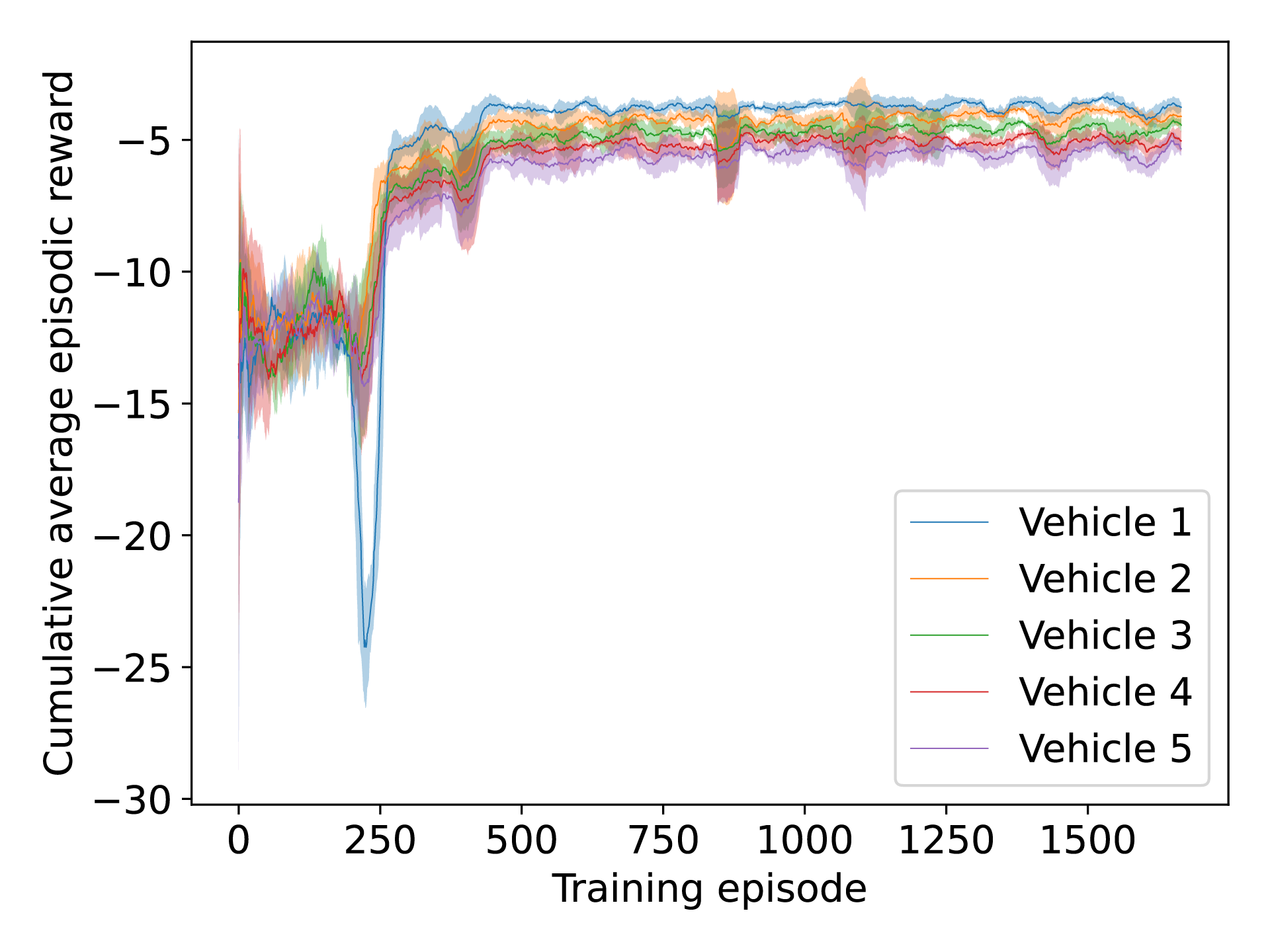}
      \caption{Intra-FRLWA: 5 Vehicles}\label{fig:reward_curve_p1_5v_intrafrlwa}
    \end{subfigure}
\caption{Average performance across 4 random seeds for 3 platoons with 3, 4 and 5 followers trained without FRL (Figures \ref{fig:reward_curve_p1_3v_nofrl},  \ref{fig:reward_curve_p1_4v_nofrl}, \ref{fig:reward_curve_p1_5v_nofrl}), and with Intra-FRLWA (Figure \ref{fig:reward_curve_p1_3v_intrafrlwa}, \ref{fig:reward_curve_p1_4v_intrafrlwa}, \ref{fig:reward_curve_p1_5v_intrafrlwa}). The shaded areas represent the standard deviation across the four seeds.}
\label{fig:vehmult-convergence}
\end{figure}

The cumulative reward is calculated over each training episode, and a moving average is computed over 40 episodes to generate Figure \ref{fig:vehmult-convergence}.  Intra-FRLWA shows favourable training performance to that of the no-FRL scenario for all platoon lengths.  In addition, the rate of convergence is increased using Intra-FRLWA versus no-FRL.  Furthermore, the shaded areas corresponding to standard deviation across the seeds are reduced significantly, indicating better stability across the seeds for Intra-FRLWA than no-FRL.  Last, the overall stability is improved as shown by the large noise reduction during training in Figure \ref{fig:reward_curve_p1_3v_intrafrlwa}, \ref{fig:reward_curve_p1_4v_intrafrlwa}, \ref{fig:reward_curve_p1_5v_intrafrlwa} when compared with no-FRL's Figure \ref{fig:reward_curve_p1_3v_nofrl},  \ref{fig:reward_curve_p1_4v_nofrl}, \ref{fig:reward_curve_p1_5v_nofrl}.

\subsubsection{Test results for one episode}
\begin{figure}[!t]
\centering
    \begin{subfigure}{0.45\textwidth}
        \centering
        \includegraphics[width=0.75\textwidth]{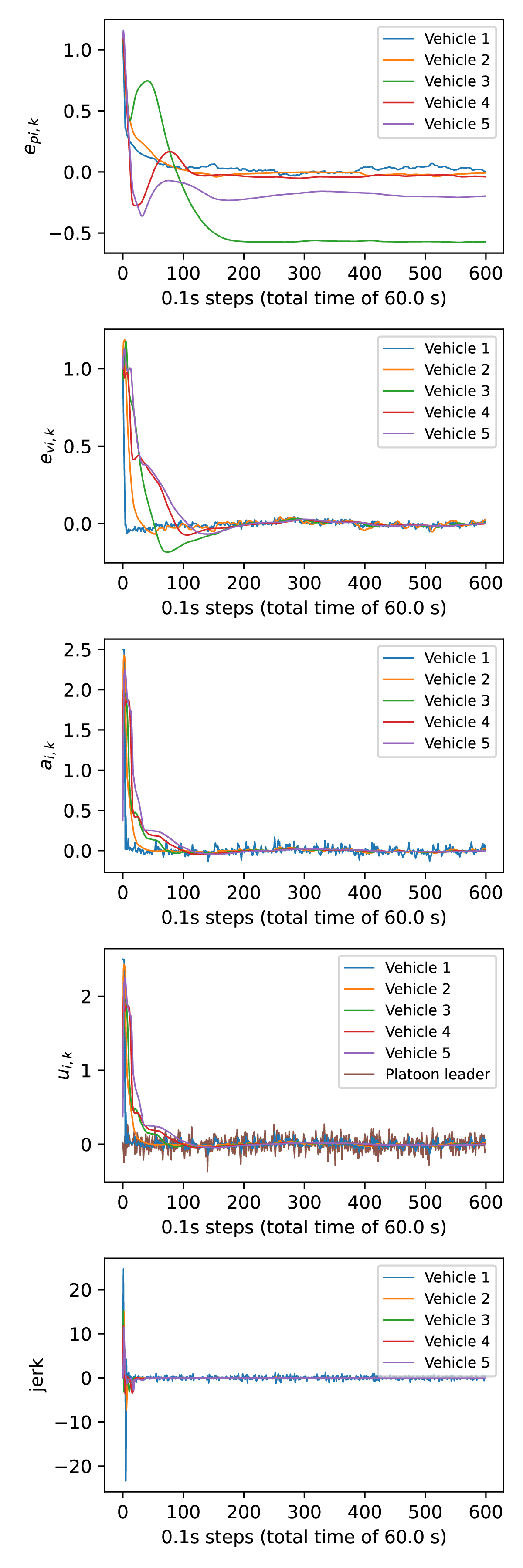}
      \caption{No-FRL}\label{fig:sim_guassian_p1_5v_nofrl}
    \end{subfigure}\hspace{\interfrlRewSpace}
    \begin{subfigure}{0.45\textwidth}
        \centering
        \includegraphics[width=0.75\textwidth]{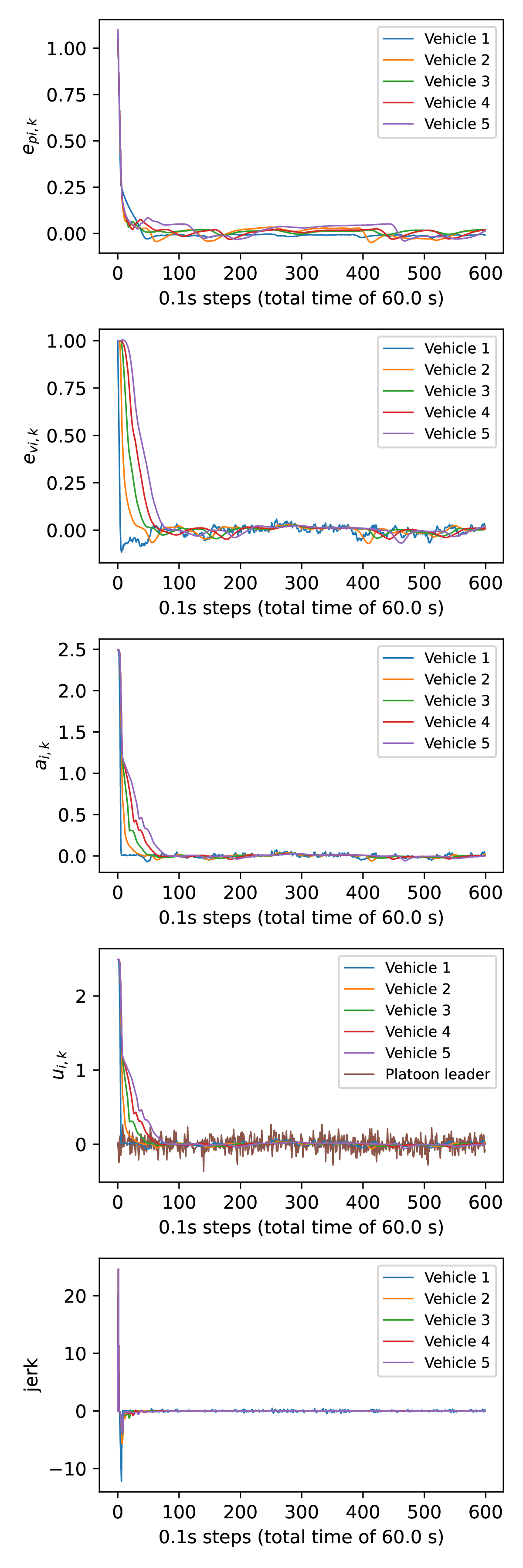}
      \caption{Intra-FRLWA}\label{fig:sim_guassian_p1_5v_intrafrlwa}
    \end{subfigure}
\caption{Results for a specific 60s test episode using the 5 vehicle 1 platoon environment trained using no-FRL (Figure \ref{fig:sim_guassian_p1_5v_nofrl}), and with Intra-FRLWA (Figure \ref{fig:sim_guassian_p1_5v_intrafrlwa}).}
\label{fig:intraFRL-vehmult-simresult}
\end{figure}

As with all previous sections, a single simulation is performed on a 60 second episode plotting the jerk along with the control input $u_{i,k}$, acceleration $a_{i,k}$, velocity error $e_{vi,k}$, and position error $e_{pi,k}$.  Figure \ref{fig:intraFRL-vehmult-simresult} showcases the ability of Intra-FRLWA to control a 5 platoon environment precisely when compared to a platoon trained without Intra-FRLWA.  The environment for Intra-FRLWA is initialized with the same values as no-FRL, just like all previous experiments: ($e_{pi} = 1.0 m$, $e_{vi}=1.0 m/s$, $a_i = 0.03 m/s^2$). Each DDPG agent trained with Intra-FRLWA quickly and precisely tracks the Gaussian random control input $u_{i,k}$ from the leader minimizing $e_{pi,k}$, $e_{vi,k}$, $a_{i,k}$ and jerk. In particular, the response for $e_{pi,k}$ and $e_{vi,k}$ in the platoon trained using Intra-FRLWA (Figure \ref{fig:sim_guassian_p1_5v_intrafrlwa}) appears to respond to the platoon leader's input quicker and in a much smoother manner than that of the no-FRL scenario (Figure \ref{fig:sim_guassian_p1_5v_nofrl}).

The large difference in performance for no-FRL versus Intra-FRL can be explained by understanding how Intra-FRLWA works. With no-FRL, each agent trains independently, and the inputs to the following vehicles are directly outputted from the predecessors. Thus, the followers farther back in the platoon take longer to train as their predecessors' outputs can be highly variable while training. As the policies of the predecessors converge, the policy of each follower can then begin to converge. This sequential convergence from predecessor to follower can be seen in Figure \ref{fig:vehmult-convergence}, where the convergence during training is slower for vehicles 4 and 5 than it is for 3, 2 and 1. Intra-FRLWA helps to resolve this challenge by allowing vehicles to average their model weights, thus distributing an aggregation of more mature predecessor parameters amongst the platoon.

\section{Conclusion}
In this paper, we have formulated an AV platooning problem and successfully applied FRL in a variety of methods to AV platooning. In addition, we proposed new approaches for applying FRL to AV platoons: Inter-FRL and Intra-FRL.  By comparing FRL performance with both gradient and weight averaging in the AV platooning scenario, it has been shown that weight averaging was the optimal aggregation method regardless of using Inter-FRL or Intra-FRL.  Furthermore, it was found that the Intra-FRLWA strategy was most advantageous for applying FRL to AV platooning.  Finally, it was proven that applying Intra-FRLWA to AV platoons up to 5 vehicles in length provided large performance advantages during and after training when compared to AV platoons that were controlled  by DDPG agents trained without FRL. These results are backed by simulations performed using models trained across four random seeds, and an additional simulation set with variable platoon sizes.  The focus of this paper was on decentralized platoon control, where each follower in the platoon trains locally with respect to their individual reward.

In the future, improvements to the system could be made by implementing weighted averaging in the FRL aggregation method. Moreover, in AV platooning, communication delays can be considered in the model to give a more concrete real life example.

\bibliography{references}{}
\bibliographystyle{IEEEtran}

\end{document}